\documentclass{article}
\usepackage{ifpdf}
     \PassOptionsToPackage{numbers, compress}{natbib}

\usepackage[final]{neurips_2019}
\usepackage[utf8]{inputenc} 
\usepackage[T1]{fontenc}    
\usepackage{hyperref}       
\usepackage{url}            
\usepackage{booktabs}       
\usepackage{amsfonts}       
\usepackage{nicefrac}       
\usepackage{microtype}      
\usepackage{pifont}
\usepackage{times}  
\usepackage{helvet} 
\usepackage{courier}  
\usepackage{subcaption}
\usepackage{graphicx} 
\usepackage{tabularx,ragged2e,booktabs,caption}
\usepackage{amsfonts}
\usepackage{bbm}
\usepackage{amssymb}
\usepackage{multirow} 
\usepackage{algorithm}
\usepackage{algorithmicx}
\usepackage{algpseudocode}
\usepackage{enumitem}
\usepackage{amsmath,bm}
\usepackage{graphicx}  
\newcommand{\infsetstyle}[1]{{\mathbb{#1}}}
\newcommand{\finsetstyle}[1]{{\mathbb{#1}}}
\newcommand{\diststyle}[1]{{\mathcal{#1}}}
\newcommand{\gp}[0]{\diststyle{GP}}
\newcommand{\tsp}[0]{{\rm T}}
\newcommand{\boolfont}[1]{{\tt #1}}
\newcommand{\ordsetstyle}[1]{{\mathfrak{#1}}}
\newcommand{\latvec}[1]{{\mbox{\boldmath $#1$}}}

\newcommand{\posquad}[0]{\bar{\infsetstyle{R}}_+^m}

\newcommand{\iverson}[1]{\mathbf{1} \left( {#1} \right)}
\newcommand{\normdist}[0]{\mathcal{N}}
\newcommand{\expect}[0]{\mathbb{E}}
\newcolumntype{C}[1]{>{\Centering}m{#1}}

\DeclareMathOperator\sgn{sgn} 
\DeclareMathOperator\argmax{argmax}

\newtheorem{def_preforder}{Definition}
\newtheorem{th_geomofS}{Theorem}
\newtheorem{cor_geomofSII}{Corollary}
\newenvironment{proof}{\paragraph{Proof:}}{\hfill$\square$}

\setlist[itemize]{leftmargin=*,label=\scalebox{.8}{\textbullet}}

\title{Multi-objective Bayesian optimisation with preferences over objectives}

\author{%
  Majid Abdolshah, Alistair Shilton, Santu Rana, Sunil Gupta, Svetha Venkatesh \\
  The Applied Artificial Intelligence Institute $(\mathrm{A^2I^2})$,\\
  Deakin University, Australia\\
  \texttt{{\{majid,alistair.shilton,santu.rana,sunil.gupta,svetha.venkatesh\}}}\\
  \texttt{{@deakin.edu.au}}
}

\begin{document}

\maketitle

\begin{abstract}
We present a  multi-objective Bayesian optimisation algorithm that allows the user to express preference-order constraints on the objectives of the type ``objective A is more important than objective B''. These preferences are defined based on the stability of the obtained solutions with respect to preferred objective functions. Rather than attempting to find a representative subset of the complete Pareto front, our algorithm selects those Pareto-optimal points that satisfy these constraints. We formulate a new acquisition function based on expected improvement in dominated hypervolume (EHI) to ensure that the subset of Pareto front satisfying the constraints is thoroughly explored. The hypervolume calculation is weighted by the probability of a point satisfying the constraints 
from a gradient Gaussian Process model. We demonstrate our algorithm on both synthetic and real-world problems.
\end{abstract}

\section{Introduction}
In many real world problems, practitioners are required to sequentially evaluate a noisy black-box and expensive to evaluate function $f$ with the goal of finding its optimum in some domain $\mathbb{X}$. Bayesian optimisation is a well-known algorithm for such problems. There are variety of studies such as hyperparameter tuning 
\cite{snoek2012practical,klein2016fast,ilievski2017efficient}, expensive multi-objective optimisation for Robotics \cite{calandra2016bayesian,calandra2014bayesian}, and experimentation optimisation in product design such as short polymer fiber materials \cite{li2017rapid}. 
\par 
Multi-objective Bayesian optimisation involves at least two conflicting, black-box, and expensive to evaluate objectives to be optimised simultaneously. Multi-objective optimisation usually assumes that all objectives are  \emph{equally important}, and solutions are found by seeking the Pareto front in the objective space \cite{deb2014multi,deb2000fast,deb2005multi}. However, in most cases, users can stipulate preferences over objectives. This information will impart on the relative importance on sections of the Pareto front. Thus using this information to preferentially sample the Pareto front will boost the efficiency of the optimiser, which is particularly advantageous when the objective functions are expensive. 
\par
In this study, preferences over objectives are stipulated based on the stability of the solutions with respect to a set of objective functions. 
As an example, there are scenarios when investment strategists are looking for Pareto optimal investment strategies that prefer stable solutions for return (objective $1$) but more diverse solutions with respect to risk (objective $2$) as they can later decide their appetite for risk. As can be inferred, the stability in one objective produces more diverse solutions for the other objectives. We believe in many real-world problems our proposed method can be useful in order to reduce the cost, and improve the safety of experimental design.
%
%
\begin{figure*}[t]
\centering
\label{fig:Mould}
\begin{subfigure}[t]{0.32\linewidth}
    \centering
    \includegraphics[width=1\linewidth]{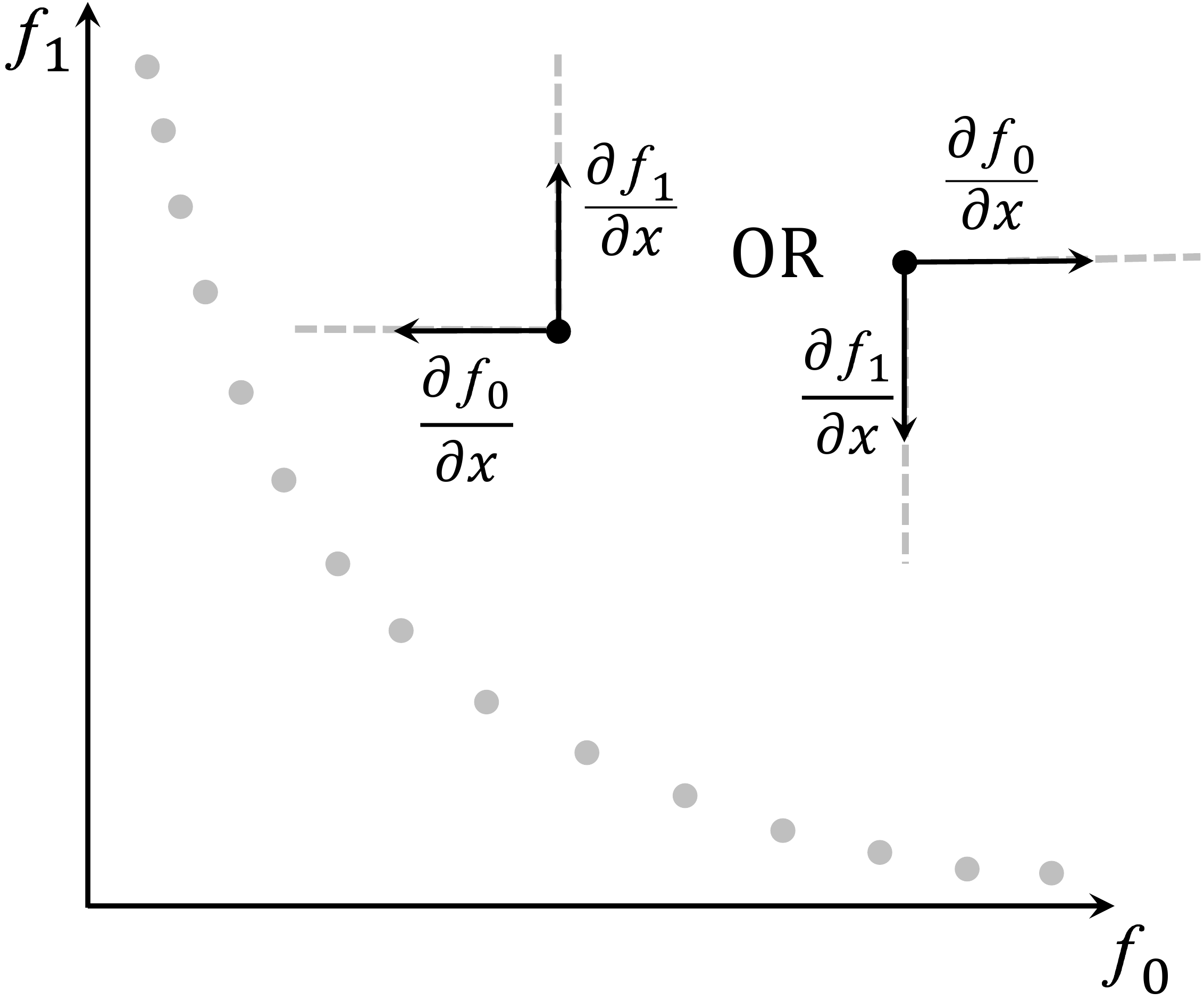}
    \caption{}
    \label{fig:int1}
\end{subfigure}
\begin{subfigure}[t]{0.32\linewidth}
    \centering
    \includegraphics[width=1\linewidth]{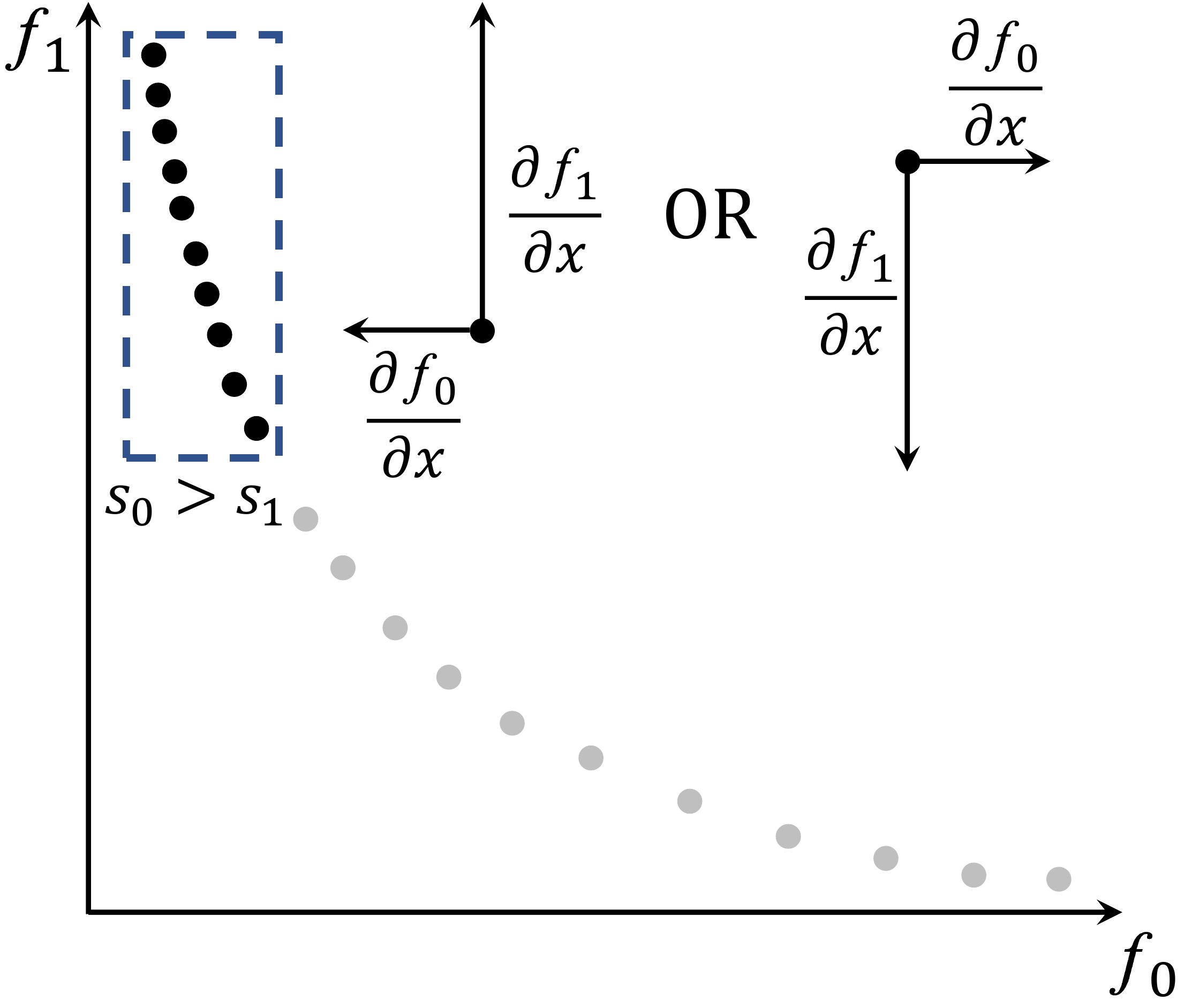}
    \caption{}
    \label{fig:int2}       
\end{subfigure}
\begin{subfigure}[t]{0.32\linewidth}
    \centering
    \includegraphics[width=1\linewidth]{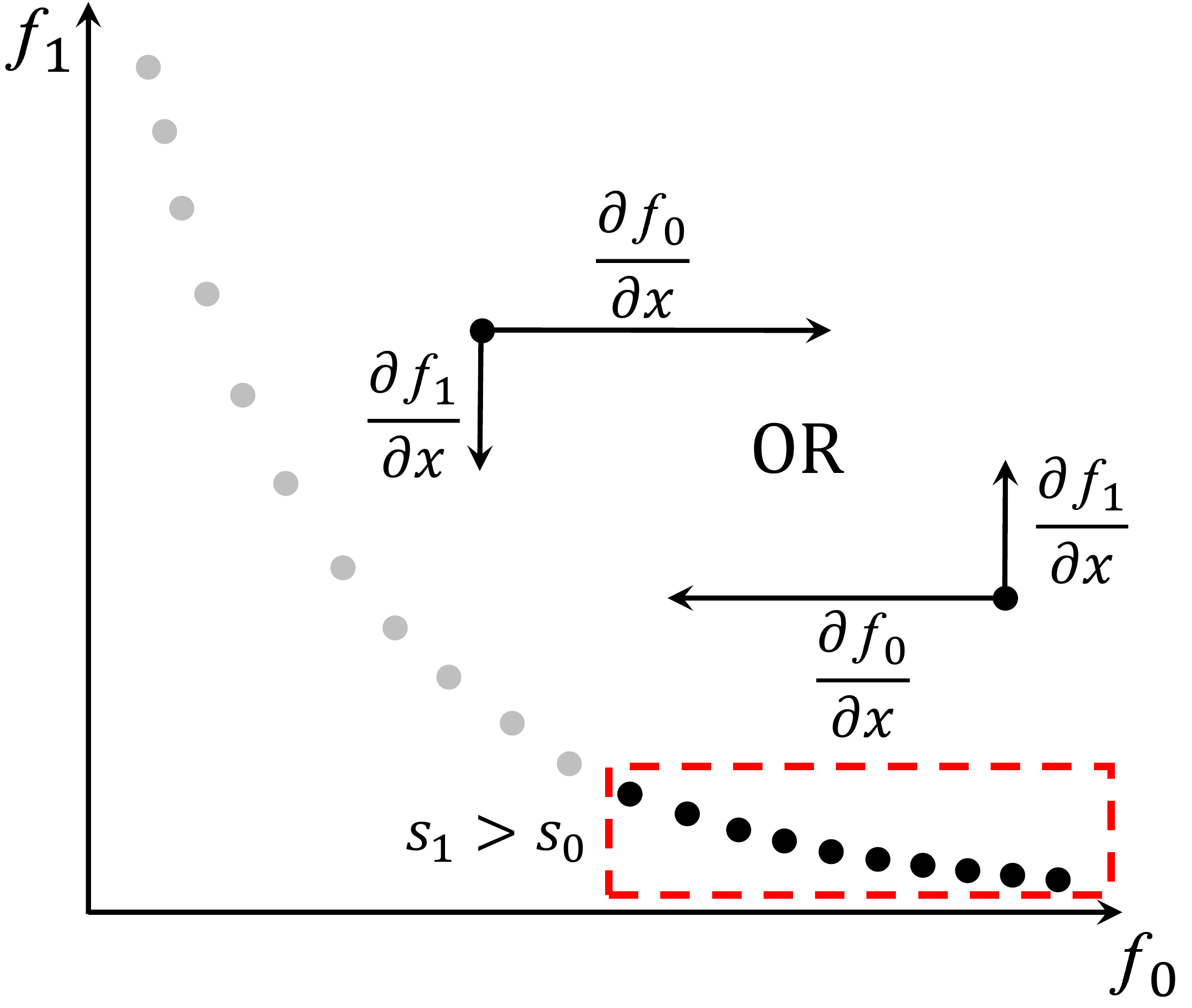}
    \caption{}
    \label{fig:int3}      
\end{subfigure}
\caption{
(a) Local Pareto optimality for $2$ objective example with $1$D design space. Local optimality implies $\frac{\partial f_0(x)}{\partial x}$ and $\frac{\partial f_1(x)}{\partial x}$ have opposite signs since the weighted sum of gradients of the objectives with respect to $x$ must be zero: ${\bf s}^{\tsp} \frac{\partial}{\partial x} {\bf f} \left( {x} \right) = 0$. In (b) we additionally require that $||\frac{\partial f_1(x)}{\partial x}|| > ||\frac{\partial f_0(x)}{\partial x}||$, so perturbation of $x$ will cause relatively more change in $f_1$ than $f_0$ - i.e. such solutions are (relatively) stable in objective $f_0$. (c) Shows the converse, namely $||\frac{\partial f_0(x)}{\partial x}|| > ||\frac{\partial f_1(x)}{\partial x}||$ favoring solutions that are (relatively) stable in objective $f_1$ and diverse in $f_0$.
}
\label{fig:patch1}
\end{figure*} 
\par
Whilst multi-objective Bayesian optimisation for sample efficient discovery of Pareto front is an established research track \cite{hernandez2016predictive,Kand2018,feliot2017bayesian,laumanns2002bayesian}, limited work has examined the incorporation of preferences. Recently, there has been a study \cite{Kand2018} 
wherein given a user specified preferred region in objective space,  the optimiser focuses its sampling to derive the Pareto front efficiently. However, such preferences are based on the assumption of having an accurate prior knowledge about objective space and the preferred region (generally a hyperbox) for Pareto front solutions.
The main contribution of this study is formulating the concept of preference-order constraints and incorporating that into a multi-objective Bayesian optimisation framework to address the unavailability of prior knowledge and boosting the performance of optimisation in such scenarios.

\par
We are formulating the preference-order constraints through ordering of derivatives and incorporating that into multi-objective optimisation using the geometry  of the constraints space whilst needing no prior information about the functions. Formally, we find a representative set of Pareto-optimal solutions to the following multi-objective optimisation problem:
\begin{equation}
 \begin{array}{l}
  \finsetstyle{D}^\star \subset \infsetstyle{X}^\star = \mathop{\argmax}\limits_{{\bf x} \in \infsetstyle{X}} {\bf f} \left( {\bf x} \right) \\
 \end{array}
 \label{eq:moo_problem}
\end{equation}
subject to {\em preference-order constraints} - that is, assuming $\mathbf{f} = [f_0, f_1,\ldots,f_m]$, $f_0$ is more important (in terms of stability) than $f_1$ and so on.
 Our algorithm aims to maximise the dominated hypervolume of the solution in a way that the solutions 
 that meet the constraints are given more \emph{weights}.
\par
To formalise the concept of preference-order constraints, we first note that a point is locally Pareto optimal if any sufficiently small perturbation of a single design parameter of that point does not simultaneously increase (or decrease) all objectives. Thus, equivalently, a point is locally Pareto optimal if we can define a set of weight vectors such that, for each design parameter, the weighted sum of gradients of the objectives with respect to that design parameter is zero (see Figure \ref{fig:int1}). Therefore, the weight vectors define the relative importance of each objective at that point. Figure \ref{fig:int2} illustrates this concept where the blue box defines the region of stability for the function $f_0$. Since in this section the magnitude of partial derivative for $f_0$ is smaller compared to that of $f_1$, the weights required to satisfy Pareto optimality would need higher weight corresponding to the gradient of $f_0$ compared to that of $f_1$ (see Figure \ref{fig:int2}). Conversely, in Figure \ref{fig:int3}, the red box highlights the section of the Pareto front where solutions have high stability in $f_1$. To obtain samples from this section of the Pareto front, we need to make the weights corresponding to the gradient of $f_0$ to be smaller to that of the $f_1$. 
\par
Our solution is based on understanding the geometry of the constraints in the weight space. We show that preference order constraints gives rise to a polyhedral proper cone in this space. We show that for the pareto-optimality condition, it necessitates the gradients of the objectives at pareto-optimal points to lie in a perpendicular cone to that polyhedral. We then quantify  the posterior probability that any point satisfies the preference-order constraints given a set of observations.  We show how these posterior probabilities may be incorporated into the EHI acquisition function \cite{Hup1} to steer the Bayesian optimiser toward Pareto optimal points that satisfy the preference-order constraint and away from those that do not. 


\section{Notation}
Sets are written $\infsetstyle{A}, \infsetstyle{B}, \infsetstyle{C}, \ldots$ 
where 
$\infsetstyle{R}_+$ is the positive reals, 
$\bar{\infsetstyle{R}}_+ = \infsetstyle{R}_+ \cup \{ 0 \}$, 
$\infsetstyle{Z}_+ = \{ 1, 2, \ldots \}$, 
and
$\infsetstyle{Z}_n = \{ 0, 1, \ldots, n-1 \}$. $|\infsetstyle{A}|$   is the cardinality of the set ${\infsetstyle{A}}$.
Tuples (ordered sets) are denoted $\ordsetstyle{A}, \ordsetstyle{B}, \ordsetstyle{C}, \ldots$. Distributions are denoted $\diststyle{A}, \diststyle{B}, \diststyle{C}, \ldots$.
column vectors are bold lower case ${\bf a}, {\bf b}, {\bf c}, \ldots$. Matrices bold upper case ${\bf A}, {\bf B}, {\bf C}, \ldots$. Element $i$ of vector ${\bf a}$ is $a_i$, 
and element $i,j$ of matrix ${\bf A}$ is $A_{i,j}$
(all indexed $i,j = 0,1,\ldots$).
The transpose is denoted ${\bf a}^{\tsp}, {\bf A}^{\tsp}$.
${\bf I}$ is the identity matrix,
${\bf 1}$ is a vector of $1$s, 
${\bf 0}$ is a vector of $0$s, 
and
${\bf e}_{i}$ is a vector $e_{(i)j} = \delta_{ij}$, 
where
$\delta_{ij}$ is the Kronecker-Delta.  
${\nabla}_{\bf x} = [ \frac{\partial}{\partial x_0} \, \frac{\partial}{\partial x_1} \, \ldots \, \frac{\partial}{\partial x_{n-1}} ]^{\tsp}$, 
$\sgn (x)$ is the sign of $x$ (where $\sgn (0) = 0$), 
and the indicator function is denoted as $\mathbbm{1} (\boolfont{A})$.

\section{Background}
\subsection{Gaussian Processes} \label{sec:gp_intro}
Let $\infsetstyle{X} \subset \infsetstyle{R}^n$ be compact.  A Gaussian process 
\cite{Ras2} $\gp ( \mu, K)$ is a distribution on the function space $f : 
\infsetstyle{X} \to \infsetstyle{R}$ defined by mean $\mu : \infsetstyle{X} \to 
\infsetstyle{R}$ (assumed zero without loss of generality) and kernel 
(covariance) $K : \infsetstyle{X} \times \infsetstyle{X} \to \infsetstyle{R}$.  
If $f ({\bf x}) \sim \gp ( 0, K ({\bf x}, {\bf x}') )$ then the posterior of 
$f$ given 
$\finsetstyle{D} = \{ ({\bf x}_{(j)}, y_{(j)}) \in \infsetstyle{R}^n \times 
\infsetstyle{R} | y_{(j)} = f ({\bf x}_{(j)}) + \epsilon, \epsilon \sim \normdist (0, 
\sigma^2), j \in \infsetstyle{Z}_N \}$, $f ({\bf x}) | \finsetstyle{D} \sim 
\normdist ( \mu_{\finsetstyle{D}} ({\bf x}), {\sigma}_{\finsetstyle{D}} ({\bf 
x}, {\bf x}'))$, where:
\begin{equation}
 \!\!\!\begin{array}{rl}
 \mu_{\finsetstyle{D}} \left( {\bf x} \right) 
   &\!\!\!= {\bf k}^{\tsp} \left( {\bf x} \right) \left( {\bf K} + {\sigma}^2 {\bf I} \right)^{-1} {\bf y} \\
 \sigma_{\finsetstyle{D}} \left( {\bf x}, {\bf x}' \right) 
 &\!\!\!= K \left( {\bf x}, {\bf x}' \right) - {\bf k}^{\tsp} \left( {\bf x} \right) \left( {\bf K} + {\sigma}^2 {\bf I} \right)^{-1} {\bf k} \left( {\bf x}' \right) 
 \end{array}\!\!\!\!\!\!\!\!
 \label{eq:gp}
\end{equation}
and  ${\bf y}, {\bf k} ({\bf x}) \in \infsetstyle{R}^{|{\infsetstyle{D}}|}$, ${\bf K} \in 
\infsetstyle{R}^{{|\infsetstyle{D}|} \times {|{ \infsetstyle{D}}|}}$, $k ({\bf x})_j = K ({\bf x}, {\bf x}_{(j)})$, $K_{jk} = K ({\bf x}_{(j)}, {\bf x}_{(k)})$.

Since differentiation is a linear operation, the derivative of a Gaussian process 
is also a Gaussian process \cite{Oha1,Ras3}. The posterior of ${\nabla}_{\bf 
x} f$ given $\finsetstyle{D}$ is ${\nabla}_{\bf x} f ({\bf x}) | 
\finsetstyle{D} \sim \normdist ({\bm \mu}_{\finsetstyle{D}}' ({\bf x}), 
{\bm \sigma}_{\finsetstyle{D}}' ({\bf x}, {\bf x}'))$, where:
\begin{align}
  {\bm \mu}_{\finsetstyle{D}}' \left( {\bf x} \right) \nonumber
   &= \left( {\nabla}_{\bf x} {\bf k}^{\tsp} \left( {\bf x} \right) \right) \left( {\bf K} + {\sigma}^2 {\bf I} \right)^{-1} {\bf y} \nonumber\\
  {\bm \sigma}_{\finsetstyle{D}}' \left( {\bf x}, {\bf x}' \right) 
  &= {\nabla}_{\bf x} {\nabla}_{{\bf x}'}^{\tsp} K \left( {\bf x}, {\bf x}' \right)
  - \left( {\nabla}_{\bf x} {\bf k}^{\tsp} \left( {\bf x} \right) \right)( {\bf K} + {\sigma}_i^2 {\bf I} )^{-1} \left( {\nabla}_{{\bf x}'} {\bf k}^{\tsp} \left( {\bf x}' \right) \right)^{\tsp}   \label{eq:gpderiv}
\end{align}
\subsection{Multi-Objective Optimisation}
A multi-objective optimisation problem has the form:
\begin{equation}
 \begin{array}{l}
  \mathop{\argmax}\limits_{{\bf x} \in \infsetstyle{X}} {\bf f} \left( {\bf x} \right) \\
 \end{array}
 \label{eq:moo_problem}
\end{equation}
where the components of ${\bf f} : \infsetstyle{X} \subset \infsetstyle{R}^n 
\to \infsetstyle{Y} \subset \infsetstyle{R}^m$ represent the $m$ distinct 
objectives $f_i : \infsetstyle{X} \to \infsetstyle{R}$.  $\infsetstyle{X}$ and 
$\infsetstyle{Y}$ are called design space and objective space, respectively.  
A Pareto-optimal solution is a point ${\bf x}^\star \in \infsetstyle{X}$ for 
which it is not possible to find another solution ${\bf x} \in \infsetstyle{X}$ 
such that $f_i ({\bf x}) > f_i ({\bf x}^\star)$ for all objectives $f_0, f_1, 
\ldots f_{m-1}$.  The set of all Pareto optimal solutions is the Pareto 
set
$
 \begin{array}{l}
  \infsetstyle{X}^\star = \left\{ \left. {\bf x}^\star \in \infsetstyle{X} \right| \nexists {\bf x} \in \infsetstyle{X} : {\bf f} \left( {\bf x} \right) \succ {\bf f} \left( {\bf x}^\star \right) \right\}
 \end{array}
$
where ${\bf y} \succ {\bf y}'$ (${\bf y}$ dominates ${\bf y}'$) means ${\bf y} 
\ne {\bf y}'$, $y_i \geq y_i'$ $\forall i$, and ${\bf y} \succeq {\bf y}'$ 
means ${\bf y} \succ {\bf y}'$ or ${\bf y} = {\bf y}'$.

Given observations $\finsetstyle{D} = \{ ({\bf x}_{(j)}, {\bf y}_{(j)}) \in 
\infsetstyle{R}^n \times \infsetstyle{R}^m | {\bf y}_{(j)} = {\bf f} ({\bf x}_{(j)}) + 
\latvec{\epsilon}, \epsilon_i \sim \normdist (0, \sigma_i^2 )\}$ of ${\bf f}$ the dominant set
$
 \begin{array}{l}
  \finsetstyle{D}^\ast = \left\{ \left. \left( {\bf x}^\ast, {\bf y}^\ast \right) \in \finsetstyle{D} \right| \nexists \left( {\bf x}, {\bf y} \right) \in \finsetstyle{D} : {\bf y} \succeq {\bf y}^\ast \right\}
 \end{array}
$
is the most optimal subset of $\finsetstyle{D}$ (in the Pareto sense).  The 
``goodness'' of $\finsetstyle{D}$ is often measured by the dominated 
hypervolume ($S$-metric, \cite{Zit1,Hub1}) with respect to some 
reference point ${\bf z} \in \infsetstyle{R}^m$:
$
 \begin{array}{l}
  S \left( \finsetstyle{D} \right) = S \left( \finsetstyle{D}^\ast \right) = \int_{{\bf y} \geq {\bf z}} \iverson{\left. \exists {\bf y}_{(i)} \in \finsetstyle{D} \right| {\bf y}_{(i)} \succeq {\bf y}} d{\bf y}.
 \end{array}
$
Thus our aim is to find the set 
$\finsetstyle{D}$ that maximises the hypervolume.  Optimised algorithms exist 
for calculating hypervolume \cite{Whi1,Shi22},
$S (\finsetstyle{D})$, which is 
typically calculated by sorting the dominant observations along each axis in 
objective space to form a grid.  Dominated hypervolume (with respect to ${\bf 
z}$) is then the sum of the hypervolumes of the dominated cells ($c_k$) - i.e. $ \begin{array}{l}
  S \left( \finsetstyle{D} \right) = \sum_k {\rm vol} \left( c_k \right).
 \end{array}
$
\subsection{Bayesian Multi-Objective Optimisation}
In the multi-objective case one typically assumes that the components of ${\bf f}$ 
are draws from independent Gaussian processes, i.e. $f_i ({\bf x}) \sim \gp (0, K_{(i)} ({\bf x}, {\bf x}'))$, and $f_i$ and $f_{i'}$ are independent $\forall 
i \ne i'$. A popular acquisition function for multi-objective Bayesian optimisation is expected hypervolume improvement (EHI).
The EHI acquisition function is defined by:
\begin{equation}
 \begin{array}{l}
  a_t \left( \left. {\bf x} \right| \finsetstyle{D} \right) = \expect_{\bf f(x)|\finsetstyle{D}} \left[ {\rm S} \left( \finsetstyle{D} \cup \left\{ \left( {\bf x}, {\bf f} \left( {\bf x} \right) \right) \right\} \right) - {\rm S} \left( \finsetstyle{D} \right) \right]
 \end{array}
 \label{eq:ehi}
\end{equation}
\cite{Shi50,Zae1} and represents the expected change in the dominated hypervolume 
 by the set of observations based on the posterior Gaussian process.

\section{Problem Formulation}
Let ${\bf f} : \infsetstyle{X} \subset \infsetstyle{R}^n \to \infsetstyle{Y} 
\subset \infsetstyle{R}^m$ be a vector of $m$ independent draws $f_i \sim \gp 
(0, K_{(i)} ({\bf x}, {\bf x}))$ from zero-mean Gaussian processes.  Assume 
that ${\bf f}$ is expensive to evaluate. Our aim is to find a representative 
set of Pareto-optimal solutions to the following multi-objective optimisation 
problem:
\begin{equation}
 \begin{array}{l}
  \finsetstyle{D}^\star \subset \infsetstyle{X}^\star = \mathop{\argmax}\limits_{{\bf x} \in \infsetstyle{X}_{\ordsetstyle{I}} \subset \infsetstyle{X}} {\bf f} \left( {\bf x} \right) \\
 \end{array}
 \label{eq:moo_problem_here}
\end{equation}

subject to {\em preference-order constraints}.  Specifically, we want to explore only that subset of solutions $\infsetstyle{X}_{\ordsetstyle{I}} 
\subset \infsetstyle{X}$ that place more {\em importance} on one objective 
$f_{i_0}$ than objective $f_{i_1}$, and so on, as specified by the (ordered) 
preference tuple $\ordsetstyle{I} = (i_0, i_1, \ldots i_Q| \{ i_0, i_1, \ldots \} 
\subset \infsetstyle{Z}_m, i_k \ne i_{k'} \forall k \ne k')$, where $Q \in \infsetstyle{Z}_m$ is the number of defined preferences over objectives. 

\subsection{Preference-Order Constraints}
Let ${\bf x}^\star \in {\rm int} (\infsetstyle{X}) \cap 
\infsetstyle{X}^\star$ be a Pareto-optimal point in the interior of 
$\infsetstyle{X}$.  Necessary (but not sufficient, local) Pareto optimality 
conditions require that, for all sufficiently small $\delta {\bf x} \in 
\infsetstyle{R}^n$, $f ({\bf x}^\star + \delta {\bf x}) \nsucc f ({\bf x})$, or, 
equivalently
$
 \begin{array}{l}
   \left(\delta {\bf x}^{\tsp} \nabla_{\bf x}\right)  {\bf f} \left( {\bf x}^\star \right)  \notin \infsetstyle{R}_+^m.
 \end{array}
$
A necessary (again not sufficient) equivalent condition is that, for each 
axis $j \in \infsetstyle{Z}_n$ in design space, sufficiently small changes in 
$x_j$ do not cause all objectives to simultaneously increase (and/or remain 
unchanged) or decrease (and/or remain unchanged). Failure of this 
condition would indicate that simply changing design parameter $x_j$ could 
improve all objectives, and hence that ${\bf x}^\star$ was not in fact Pareto 
optimal.  In summary, local Pareto optimality requires that $\forall j \in \infsetstyle{Z}_n$ there exists ${\bf 
s}_{(j)} \in \posquad \backslash \{ {\bf 0} \}$ such that:
\begin{equation}
 \begin{array}{l}
  {\bf s}_{(j)}^{\tsp} \frac{\partial}{\partial x_j} {\bf f} \left( {\bf x} \right) = 0
 \end{array}
 \label{eq:our_pref_cond}
\end{equation}
It is important to note that this is not the same as the optimality conditions 
that may be derived from linear scalarisation, as the optimality conditions 
that arrise from linear scalarisation additionally require that ${\bf s}_{(0)} 
= {\bf s}_{(1)} = \ldots = {\bf s}_{(n-1)}$.  Moreover (\ref{eq:our_pref_cond}) 
applies to all Pareto-optimal points, whereas linear scalarisation optimisation 
conditions fail for Pareto points on non-convex regions \cite{van2013scalarized}. 

\begin{def_preforder}[Preference-Order Constraints]
 Let $\ordsetstyle{I} = (i_0, i_1, \ldots i_Q| \{ i_0, i_1, \ldots \} \subset 
 \infsetstyle{Z}_m, i_k \ne i_{k'} \forall k \ne k')$ be an (ordered) 
 preference tuple.  A vector ${\bf x} \in \infsetstyle{X}$ satisfies the 
 associated preference-order constraint if $\exists {\bf s}_{(0)}, {\bf 
 s}_{(1)}, \ldots, {\bf s}_{(n-1)} \in \infsetstyle{S}_{\ordsetstyle{I}}$ such 
 that:
\[
  \begin{array}{l}
   {\bf s}_{(j)}^{\tsp} \frac{\partial}{\partial x_j} {\bf f} \left( {\bf x} \right) = 0 \; \forall j \in \infsetstyle{Z}_n
  \end{array}
\]
 where $
  \begin{array}{l}
   \infsetstyle{S}_{\ordsetstyle{I}} \triangleq \left\{ \left. {\bf s} \in \posquad \backslash \left\{ {\bf 0} \right\} \right| s_{i_0} \geq s_{i_1} \geq s_{i_2} \geq \ldots \right\}.
  \end{array}
$
 Further we define $\infsetstyle{X}_{\ordsetstyle{I}}$ to be the set of all 
 ${\bf x} \in \infsetstyle{X}$ satisfying the preference-order constraint.  
 Equivalently:

 \[
  \begin{array}{l}
   \infsetstyle{X}_{\ordsetstyle{I}} = \{ {\bf x} \in \infsetstyle{X}_{\ordsetstyle{I}} | \frac{\partial}{\partial x_j} {\bf f} \left( {\bf x} \right) \in \infsetstyle{S}_{\ordsetstyle{I}}^{\perp} \; \forall j \in \infsetstyle{Z}_n\}
  \end{array}
 \]
 where $
  \begin{array}{l}
   \infsetstyle{S}_{\ordsetstyle{I}}^\perp \triangleq \left\{ \left. {\bf x} \in \infsetstyle{X} \right| \exists {\bf s} \in \infsetstyle{S}_{\ordsetstyle{I}}, {\bf s}^{\tsp} {\bf x} = 0 \right\}.
  \end{array}
$
 \label{def:def_preforder}
\end{def_preforder}
It is noteworthy to mention that (\ref{eq:our_pref_cond}) and Definition \ref{def:def_preforder} are the key for calculating the compliance of a recommended solution with the preference-order constraints. Having defined preference-order constraints we then calculate the posterior probability that ${\bf x} \in \infsetstyle{X}_{\ordsetstyle{I}}$, and showing how these posterior probabilities may be incorporated into the EHI acquisition function to steer the Bayesian optimiser toward Pareto optimal points that satisfy the preference-order constraint. 
Before proceeding, however, it is necessary to briefly consider the geometry of $\infsetstyle{S}_{\ordsetstyle{I}}$ and $\infsetstyle{S}_{\ordsetstyle{I}}^\perp$.

\subsection{The Geometry of $\infsetstyle{S}_{\ordsetstyle{I}}$ and $\infsetstyle{S}_{\ordsetstyle{I}}^\perp$}
In the following we assume, w.l.o.g, that the preference-order constraints follows the order of indices in objective functions (reorder, otherwise), and that there is at least one constraint. 
\par
We now define the preference-order constraints by assumption $\ordsetstyle{I} = (0, 1, \ldots,Q| Q \in \infsetstyle{Z}_m \textbackslash \{0\})$, where $Q > 0$.
This defines the sets $\infsetstyle{S}_{\ordsetstyle{I}}$ and 
$\infsetstyle{S}_{\ordsetstyle{I}}^\perp$, which in turn define the constraints 
that must be met by the gradients of ${\bf f} ({\bf x})$ - either $\exists {\bf 
s}_{(0)}, {\bf s}_{(1)}, \ldots, {\bf s}_{(n-1)} \in 
\infsetstyle{S}_{\ordsetstyle{I}}$ such that $
 \begin{array}{l}
  {\bf s}_{(j)}^{\tsp} \frac{\partial}{\partial x_j} {\bf f} \left( {\bf x} \right) = 0 \; \forall j \in \infsetstyle{Z}_n
 \end{array}
$
or, equivalently $
 \begin{array}{l}
  \frac{\partial}{\partial x_j} {\bf f} \left( {\bf x} \right) \in \infsetstyle{S}_{\ordsetstyle{I}}^\perp \; \forall j \in \infsetstyle{Z}_n.
 \end{array}
$
Next, Theorem \ref{th:th_geomofS} defines the representation of $\infsetstyle{S}_{\ordsetstyle{I}}$.

\begin{figure}
\centering
\includegraphics[width=0.4\textwidth]{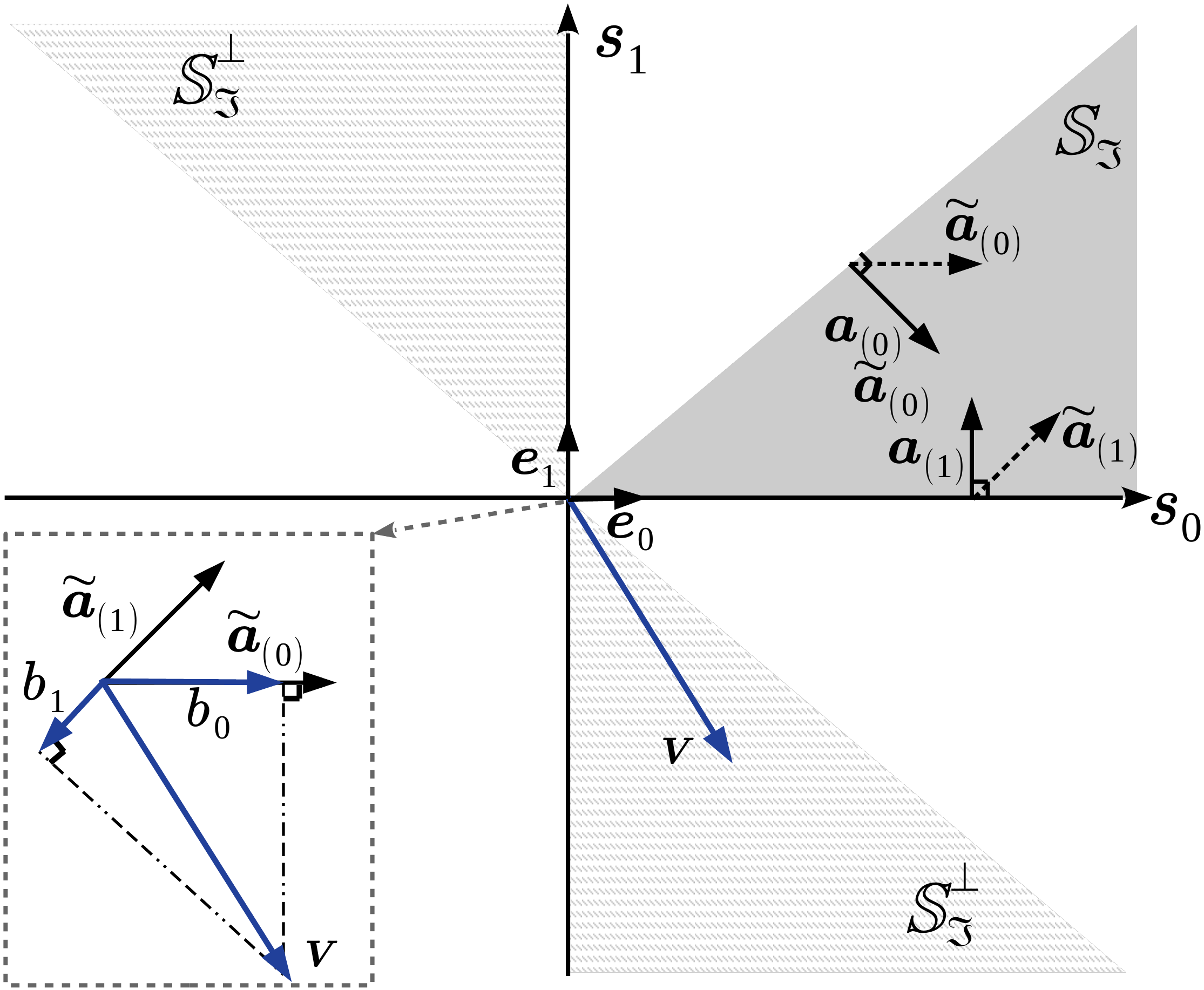}
\caption{
Illustration of $\infsetstyle{S}_{\ordsetstyle{I}}^\perp, \infsetstyle{S}_{\ordsetstyle{I}}$ and the vectors ${\bf a}_{(0)}, {\bf a}_{(1)}$ for a $2$D case where $\ordsetstyle{I} = (0, 1)$, so $s_0 > s_1$, $\infsetstyle{S}_{\ordsetstyle{I}}$ is a proper cone representing  the preference-order constraints; $\infsetstyle{S}_{\ordsetstyle{I}}^\perp$ is the union of two sub-spaces. $\mathbf{v} \in \infsetstyle{S}_{\ordsetstyle{I}}^\perp$ implies a solution complying with preference-order constraints. ${b}_0$ and ${b}_1$  are the projection of $\mathbf{v}$ over $\tilde{{\bf a}}_{(0)}$ and $\tilde{{\bf a}}_{(1)}$. In order to satisfy $\mathbf{v} \in \infsetstyle{S}_{\ordsetstyle{I}}^\perp$, it is necessary that $ \exists {\bf s} \in \infsetstyle{S}_{\ordsetstyle{I}}\ s.t.\ {\bf v}^T{\bf s} = 0$ or equivalently ${\bf v = 0}$ or ${b}_0 = \tilde{{\bf a}}_{(0)}^T {\bf v}$ and ${b}_1 = \tilde{{\bf a}}_{(1)}^T {\bf v}$ have different signs.
}
\label{fig:E3}
\end{figure}

\begin{th_geomofS} 
 Let $\ordsetstyle{I} = (0, 1, \ldots,Q| Q \in \infsetstyle{Z}_m \textbackslash \{0\})$ be an (ordered) 
 preference tuple.  Define $\infsetstyle{S}_{\ordsetstyle{I}}$ as per 
 definition \ref{def:def_preforder}.  Then $\infsetstyle{S}_{\ordsetstyle{I}}$ 
 is a polyhedral (finitely-generated) proper cone (excluding the origin) that 
 may be represented using either a polyhedral representation:
 \begin{equation}
  \begin{array}{l}
   \infsetstyle{S}_{\ordsetstyle{I}} = \left\{ \left. {\bf s} \in \infsetstyle{R}^m \right| {\bf a}_{(i)}^{\tsp} {\bf s} \geq 0 \forall i \in \infsetstyle{Z}_m \right\} \backslash \left\{ {\bf 0} \right\} \\
  \end{array}
  \label{eq:defSpoly}
 \end{equation} 
 or a generative representation:
 \begin{equation}
  \begin{array}{rl}
   \infsetstyle{S}_{\ordsetstyle{I}} &\!\!\!\!= \Big\{ \mathop{\sum}\limits_{i \in \infsetstyle{Z}_m} c_i \tilde{\bf a}_{(i)} \left|\; {\bf c} \in \posquad \right. \Big\} \backslash \left\{ {\bf 0} \right\} \\
  \end{array}
  \label{eq:defSfingen}
 \end{equation} 
 where $\forall i \in \infsetstyle{Z}_m$:
 \[
  \begin{array}{l}
  \begin{array}{rl}
   {\bf a}_{(i)} &\!\!\!\!\!= \left\{ \!\!\begin{array}{ll} 
    \frac{1}{\sqrt{2}} \left( {{\bf e}_{i} - {\bf e}_{i+1}} \right) & \!\!\mbox{if } i \in \infsetstyle{Z}_Q \\ 
    {\bf e}_{i}                                                             & \!\!\mbox{otherwise} \\ 
   \end{array} \!\!\right. \\
   \tilde{\bf a}_{(i)} &\!\!\!\!\!= \left\{ \!\!\begin{array}{ll} 
    \frac{1}{\sqrt{i+1}} \mathop{\sum}\limits_{ l  \in \infsetstyle{Z}_{i+1}} {\bf e}_{l} & \!\!\mbox{if } i \in \infsetstyle{Z}_{Q+1} \\ 
    {\bf e}_{i}                                                                           & \!\!\mbox{otherwise} \\ 
   \end{array} \!\!\right. \\
  \end{array}
  \end{array}
 \]
 and ${\bf e}_{0}, {\bf e}_{1}, \ldots, {\bf e}_{m-1}$ are the Euclidean 
 basis of $\infsetstyle{R}^m$.
 \label{th:th_geomofS}
\end{th_geomofS}
Proof of Theorem \ref{th:th_geomofS} is available in the supplementary material. 
To test if a point satisfies this requirement we need to understand the 
geometry of the set $\infsetstyle{S}_{\ordsetstyle{I}}$.  The Theorem \ref{th:th_geomofS} shows that $\infsetstyle{S}_{\ordsetstyle{I}} 
\cup \{ {\bf 0} \}$ is a polyhedral (finitely generated) proper cone, represented either in terms of half-space constraints (polyhedral form) or as a positive span of 
 extreme directions (generative representation). The geometrical intuition for this is given in Figure \ref{fig:E3} for a simple, 
$2$-objective case with a single preference order constraint. 
\begin{minipage}{0.48\textwidth}
\begin{algorithm}[H]
\caption{Test if ${\bf v} \in \infsetstyle{S}_{\ordsetstyle{I}}^\perp$.}
\label{alg:sperp_test}
\begin{algorithmic}
 \State \textbf{Input:\ \ } Preference tuple  $\ordsetstyle{I}$
 \State Test vector ${\bf v} \in \infsetstyle{R}^m$.
 \State \textbf{Output:\ \ }$\mathbbm{1} ( {\bf v} \in \infsetstyle{S}_{\ordsetstyle{I}}^\perp )$.
 \State {\em $//$ Calculate $\mathbf{1} ( {\bf v} \in \infsetstyle{S}_{\ordsetstyle{I}}^\perp )$.}
 \State Let $b_j = \tilde{\bf a}_{(j)}^{\tsp} {\bf v}$ $\forall j \in \infsetstyle{Z}_m$.
 \State if $\exists i \ne k \in \infsetstyle{Z}_m : \sgn (b_i) \ne \sgn (b_k)$ return ${\tt TRUE}$
 \State elseif ${\bf b} = {\bf 0}$ return ${\tt TRUE}$
 \State else return ${\tt FALSE}$.
 \end{algorithmic}
\end{algorithm} 
\end{minipage}
\hfill
\begin{minipage}{0.48\textwidth}
\begin{algorithm}[H]
\caption{Preference-Order Constrained Bayesian Optimisation (MOBO-PC).}
\label{alg:modded_bbo}
\begin{algorithmic}
\State \textbf{Input:\ \ } preference-order tuple $\ordsetstyle{I}$.\\
 Observations $\finsetstyle{D} = \{ ({\bf x}_{(i)}, {\bf y}_{(i)}) \in \infsetstyle{X} \times \infsetstyle{Y} \}$.
 \For{$t=0,1,\ldots,T-1$}
 \State Select the test point:  \\
 $\ \ \ \ \ \ \ {\bf x} = \mathop{\argmax}\limits_{{\bf x} \in 
        \infsetstyle{X}} a_t^{\rm PEHI} ({\bf x}|\finsetstyle{D}_t)$.\\
 \State ($a_t^{\rm PEHI}$ is evaluated using algorithm \ref{alg:acquisition}).
 \State Perform Experiment ${\bf y} = {\bf f} ( {\bf x} ) + \latvec{\epsilon}$.
 \State Update $\finsetstyle{D}_{t+1} := \finsetstyle{D}_t \cup \{ ({\bf x}, {\bf y}) \}$.
 \EndFor
\end{algorithmic}
\end{algorithm}
\end{minipage}
\hfill
\begin{minipage}{0.48\textwidth}
\begin{algorithm}[H]
\caption{Calculate $\Pr ( {\bf x} \in \infsetstyle{X}_{\ordsetstyle{I}} | \finsetstyle{D} )$.}
\label{alg:probgen}
\begin{algorithmic}
  \State \textbf{Input:\ \ } Observations $\finsetstyle{D} = \{ ({\bf x}_{(i)}, {\bf y}_{(i)}) \in \infsetstyle{X} \times \infsetstyle{Y} \}$.
 \State Number of Monte Carlo samples $R$.
 \State Test vector ${\bf x} \in \infsetstyle{X}$.
 \State \textbf{Output:\ \ } $\Pr ( {\bf x} \in \infsetstyle{X}_{\ordsetstyle{I}} | \finsetstyle{D} )$.

 \State Let $q = 0$.
 \For{$k=0,1,\ldots,R-1$}
 \State {\em $//{Construct\ samples}$ \\
 $\ \ \ \ \ \ {\bf v}_{(0)}, {\bf v}_{(1)}, \ldots, {\bf v}_{(n-1)} \in \infsetstyle{R}^m$.}
 \State Let ${\bf v}_{(j)} = {\bf 0}$ $\forall j \in \infsetstyle{Z}_n$.
 \For{$i = 0,1,\ldots,m-1$}
 \State Sample ${\bf u} \sim \normdist ({\bm \mu}_{\finsetstyle{D} i}' ({\bf x}), 
       {\bm \sigma}_{\finsetstyle{D} i}' ({\bf x},{\bf x}))$ \\
       \ \ \ \ \ \ \ \ \ \  \ \ (see (\ref{eq:gpderiv})).
 \State Let $[ v_{(0)i}, v_{(1)i}, \ldots, v_{(n-1)i} ] := {\bf u}^{\tsp}$.
 \EndFor

 \State {\em $// Test\ if$ ${\bf v}_{(j)} \in \infsetstyle{S}_{\ordsetstyle{I}}^\perp$ $\forall j \in \infsetstyle{Z}_n$.}
 \State Let $q := q + \mathop{\prod}\limits_{j \in \infsetstyle{Z}_n} \mathbbm{1} ( {\bf v}_{(j)} \in \infsetstyle{S}_{\ordsetstyle{I}}^\perp )$ (see algo \\
 \ \ \ \ \ \ rithm \ref{alg:sperp_test}).
 \EndFor
 \State Return $\frac{q}{R}$.
 \end{algorithmic}
 \label{Alg:Patch}
\end{algorithm}
\end{minipage}
\hfill
\begin{minipage}{0.48\textwidth}
\begin{algorithm}[H]
\caption{Calculate $a_t^{\rm PEHI} ( {\bf x} | \finsetstyle{D} )$.}
\label{alg:acquisition}
\begin{algorithmic}
\State \textbf{Input:\ \ } Observations $\finsetstyle{D} = \{ ({\bf x}_{(i)}, {\bf y}_{(i)}) \in \infsetstyle{X} \times \infsetstyle{Y} \}$.
 \State Number of Monte Carlo samples $\tilde{R}$.
 \State Test vector ${\bf x} \in \infsetstyle{X}$.
 \State \textbf{Output:\ \ } $a_t^{\rm PEHI} ( {\bf x} | \finsetstyle{D} )$.
 \State Using algorithm \ref{alg:probgen}, calculate:\\
$
         \begin{array}{rl}
          s_x     &\!\!\!= \Pr \left( \left. {\bf x} \in \infsetstyle{X}_{\ordsetstyle{I}} \right| \finsetstyle{D} \right) \\
          s_{(j)} &\!\!\!= \Pr \left( \left. {\bf x}_{(j)} \in \infsetstyle{X}_{\ordsetstyle{I}} \right| \finsetstyle{D} \right) \; \forall \left( {\bf x}_{(j)}, {\bf y}_{(j)} \right) \in \finsetstyle{D} \\
         \end{array}
$
 \State Let $q = 0$.
 \For{$k=0,1,\ldots,\tilde{R}-1$}
 \State Sample $y_i \sim \normdist (\mu_{\finsetstyle{D} i} ({\bf x}), \sigma_{\finsetstyle{D} i}({\bf x})) )$ $\forall i \in$ \\
 $\ \ \ \ \ \ \ \finsetstyle{Z}_m$ (see (\ref{eq:gp})).
 \State Construct cells $c_0, c_1, \ldots$ from $\finsetstyle{D} \cup$\\
 $\ \ \ \ \ \{ ({\bf  x}, {\bf y}) \}$ by sorting along each axis in \\
\ \ \ \ \ objective space to form a grid.
  \State Calculate:\\
$
         \begin{array}{l}
          \quad \quad q = q + \\
			\quad   s_x \mathop{\sum}\limits_{k : {\bf y} \succeq \tilde{\bf y}_{c_k}} {\rm vol} \left( c_k \right) 
          \mathop{\prod}\limits_{j \in \infsetstyle{Z}_N : {\bf y}_{(j)} \succeq \tilde{\bf y}_{c_k}} \left( 1 - s_{(j)} \right)
         \end{array}
         \label{eq:ehi_here_cell}
$
 \EndFor
 \State Return ${q}/{\tilde{R}}$.
 \end{algorithmic}
\end{algorithm} 
\end{minipage}

The subsequent corollary allows us to construct a simple algorithm (algorithm 
\ref{alg:sperp_test}) to test if a vector ${\bf v}$ lies in the set 
$\infsetstyle{S}_{\ordsetstyle{I}}^\perp$.  We will use this algorithm to test 
if $\frac{\partial}{\partial x_j} {\bf f} ({\bf x}) \in 
\infsetstyle{S}_{\ordsetstyle{I}}^\perp \; \forall j \in \infsetstyle{Z}_n$ - 
that is, if ${\bf x}$ satisfies the preference-order constraints. The proof of corollary \ref{cor:cor_geomofSII} is available in the supplementary material.
\begin{cor_geomofSII} 
 Let $\ordsetstyle{I} = (0, 1, \ldots,Q| Q \in \infsetstyle{Z}_m \textbackslash \{0\})$ be an (ordered) 
 preference tuple. Define $\infsetstyle{S}_{\ordsetstyle{I}}^\perp$ as per 
 definition \ref{def:def_preforder}.  Using the notation of Theorem 
 \ref{th:th_geomofS},  ${\bf v} \in \infsetstyle{S}_{\ordsetstyle{I}}^\perp$ if 
 and only if ${\bf v} = {\bf 0}$ or $\exists i \ne k \in \infsetstyle{Z}_m$ such 
 that $\sgn ( \tilde{\bf a}_{(i)}^{\tsp} {\bf v} ) \ne \sgn ( \tilde{\bf 
 a}_{(k)}^{\tsp} {\bf v} )$, where $\sgn (0) = 0$.
 \label{cor:cor_geomofSII}
\end{cor_geomofSII}

\section{Preference Constrained Bayesian Optimisation}
In this section we do two things.  First, we show how the Gaussian process models of the 
objectives $f_i$ (and their derivatives) may be used to calculate the posterior 
probability that ${\bf x} \in \infsetstyle{X}_\ordsetstyle{I}$ defined by 
$\ordsetstyle{I} = (0, 1, \ldots,Q| Q \in \infsetstyle{Z}_m \textbackslash \{0\})$. Second, we show how the EHI acquisition function may be modified and calculated to incorporate these probabilities and hence only reward points that satisfy the 
preference-order conditions. Finally, we give our algorithm using this acquisition function.

\subsection{Calculating Posterior Probabilities}
Given that $f_i \sim \gp (0, K_{(i)} ({\bf x},{\bf x}))$ are draws from independent Gaussian processes,  and given observations $\finsetstyle{D}$, we
wish to calculate the posterior probability that ${\bf x}$ $\in \infsetstyle{X}_\ordsetstyle{I}$ - i.e.:$
 \begin{array}{l}
  \Pr \left( \left. {\bf x} \in \infsetstyle{X}_{\ordsetstyle{I}} \right| \finsetstyle{D} \right) = \Pr \left( \frac{\partial}{\partial x_j} {\bf f} \left( {\bf x} \right) \in \infsetstyle{S}_{\ordsetstyle{I}}^{\perp} \; \forall j \in \infsetstyle{Z}_n \right).
 \end{array}
$
As $f_i \sim \gp (0, K_{(i)} ({\bf x},{\bf x}))$ it follows that ${\nabla}_{\bf 
x} f_i ({\bf x}) | \finsetstyle{D} \sim \normdist_i \triangleq \normdist ({\bm 
\mu}_{\finsetstyle{D} i}' ({\bf x}), {\bm \sigma}_{\finsetstyle{D} i}' 
({\bf x}, {\bf x}'))$, as defined by (\ref{eq:gpderiv}).  Hence:
\[
 \begin{array}{l}
  \Pr \left( \left. {\bf x} \in \infsetstyle{X}_{\ordsetstyle{I}} \right| \finsetstyle{D} \right) 
  = \Pr \left( \!\!\!\begin{array}{r} \; \\ \!{\bf v}_{(j)} \in \infsetstyle{S}_{\ordsetstyle{I}}^\perp \\ \forall j \in \infsetstyle{Z}_n \\ \end{array}\!\!\! \left| \left[ \begin{array}{c} v_{(0)i} \\ v_{(1)i} \\ \vdots \\ v_{(n-1)i} \\ \end{array} \right] \!\!\!\begin{array}{l} \; \\ \sim \normdist_i \\ \;\forall i \in \infsetstyle{Z}_m\!\! \end{array}\!\!\! \right. \right)
 \end{array}
\]
where ${\bf v}\sim P(\nabla_{\bf x} {\bf f}|\finsetstyle{D})$. We estimate it using Monte-Carlo \cite{del2006sequential} sampling as per algorithm \ref{alg:probgen}.

\subsection{Preference-Order Constrained Bayesian Optimisation Algorithm (MOBO-PC)}
Our complete Bayesian optimisation algorithm with Preference-order constraints 
is given in algorithm \ref{alg:modded_bbo}.  The acquisition function introduced in 
this algorithm 
gives higher importance to points satisfying the preference-order 
constraints.
Unlike standard EHI, we take expectation 
over both the expected experimental outcomes $f_i ({\bf x}) \sim \normdist 
(\mu_{\finsetstyle{D}i} ({\bf x}), \sigma_{\finsetstyle{D}i} ({\bf x}, {\bf 
x}))$, $\forall i \in \infsetstyle{Z}_m$, and the probability that points ${\bf x}_{(i)} \in \infsetstyle{X}_\ordsetstyle{I}$ and 
${\bf x} \in \infsetstyle{X}_\ordsetstyle{I}$ satisfy the preference-order constraints.  We define our 
preference-based EHI acquisition function as:
\begin{equation}
 \begin{array}{c}
  a_t^{\rm PEHI} \left( \left. {\bf x} \right| \finsetstyle{D} \right) = \expect \left[ \left. {\rm S}_{\ordsetstyle{I}} \left( \finsetstyle{D} \cup \left\{ \left( {\bf x}, {\bf f} \left( {\bf x} \right) \right) \right\} \right) - {\rm S}_{\ordsetstyle{I}} \left( \finsetstyle{D} \right) \right| \finsetstyle{D} \right]
 \end{array}
 \label{eq:ehi_here}
\end{equation}
where $S_{\ordsetstyle{I}} (\finsetstyle{D})$ is the hypervolume dominated by 
the observations $({\bf x}, {\bf y}) \in \finsetstyle{D}$ satisfying the 
preference-order constraints.  The calculation of $S_{\ordsetstyle{I}} 
(\finsetstyle{D})$  is illustrated in the supplementary material. 
The expectation of $S_{\ordsetstyle{I}} (\finsetstyle{D})$ given $\finsetstyle{D}$ is:\\\ \\
$
 \begin{array}{l}
  \!\!\!\begin{array}{rl}
   \expect \left[ \left. S_{\ordsetstyle{I}} \left( \finsetstyle{D} \right) \right| \finsetstyle{D} \right]
   =&\!\!\! \!\!\mathop{\sum}\limits_k \!{\rm vol} \left( c_k \right) \Pr ( \left. \exists \left( {\bf x}, {\bf y} \right) \!\in \!\finsetstyle{D} \right| {\bf y} \!\succeq\! \tilde{\bf y}_{c_k} \wedge \ldots {\bf x} \in \infsetstyle{X}_{\ordsetstyle{I}} ) \ldots \\
 \end{array} \\
  = \mathop{\sum}\limits_k {\rm vol} \left( c_k \right) ( 1 - \mathop{\prod}\limits_{({\bf x},{\bf y}) \in \finsetstyle{D} : {\bf y} \succeq \tilde{\bf y}_{c_k}} \left( 1 - \Pr \left( \left. {\bf x} \in \infsetstyle{X}_{\ordsetstyle{I}} \right| \finsetstyle{D} \right) \right) ) \\
 \end{array}
$
\\where $\tilde{\bf y}_{c_k}$ is the dominant corner of cell $c_k$, ${\rm vol}(c_k)$ is the hypervolume of cell $c_k$, 
and the cells $c_k$ are constructed by sorting $\finsetstyle{D}$ along each 
axis in objective space.
 The posterior probabilities $\Pr 
({\bf x} \in \infsetstyle{X}_{\ordsetstyle{I}} | \finsetstyle{D} )$ are 
calculated using algorithm \ref{alg:probgen}.  It follows that:
\[
 \begin{array}{l}
 \vspace{-0.2cm}
  a_t^{\rm PEHI} \left( \left. {\bf x} \right| \finsetstyle{D} \right)
  = \  
  \Pr \left( \left. {\bf x} \in \infsetstyle{X}_{\ordsetstyle{I}} \right| \finsetstyle{D} \right) \expect \Big[ \!\!\mathop{\sum}\limits_{\ \ k : {\bf y} \succeq \tilde{\bf y}_{c_k}} \!\!\!\!{\rm vol} \left( c_k \right) 
   \!\!\!\mathop{\prod}\limits_{j \in \infsetstyle{Z}_N : {\bf y}_{(j)} \succeq \tilde{\bf y}_{c_k}} \!\!\!\left( 1 - \Pr \left( \left. {\bf x}_{(j)} \in \infsetstyle{X}_{\ordsetstyle{I}} \right| \finsetstyle{D} \right) \right) \Big| y_i \sim \ldots \\
   \quad\quad\quad\quad\quad\quad\quad\quad\quad\quad\quad\quad\quad\quad\quad\quad\quad\quad\quad\quad\quad\quad\quad\quad\quad\quad \normdist \left( \mu_{\finsetstyle{D} i} \left( {\bf x} \right), \sigma_{\finsetstyle{D} i} \left( {\bf x} \right) \right) \; \forall i \in \infsetstyle{Z}_m \Big] \\
 \end{array}
 \label{eq:ehi_here_cell}
\]
where the cells $c_k$ are constructed using the 
set $\finsetstyle{D} \cup \{ ({\bf x}, {\bf y}) \}$ by sorting along the axis in objective space.
We estimate this 
acquisition function using Monte-Carlo simulation shown in algorithm \ref{alg:acquisition}.
\section{Experiments}
We conduct a series of experiments to test the empirical performance of our proposed method MOBO-PC and compare with other strategies.  These experiments including synthetic data as well as optimizing  the  hyper-parameters of a feed-forward neural network. For Gaussian process, we use maximum likelihood estimation for setting hyperparameters \cite{rasmussen2004gaussian}.

\subsection{Baselines} 
To the best of our knowledge there are no studies aiming to solve our proposed problem, however we are using PESMO, SMSego, SUR, ParEGO and EHI \cite{hernandez2016predictive,ponweiser2008multiobjective,picheny2015multiobjective,knowles2006parego,emmerich2008computation} to confirm the validity of the obtained Pareto front solutions. The obtained Pareto front must be in the ground-truth whilst also satisfying the preference-order constraints. We compare our results with MOBO-RS \cite{Kand2018} by suitably specifying bounding boxes in the objective space that can replicate a preference-order constraint.
%
\begin{figure}
\centering
\label{fig:Mould}
\begin{subfigure}[t]{0.3\linewidth}
    \centering
    \includegraphics[width=1\linewidth]{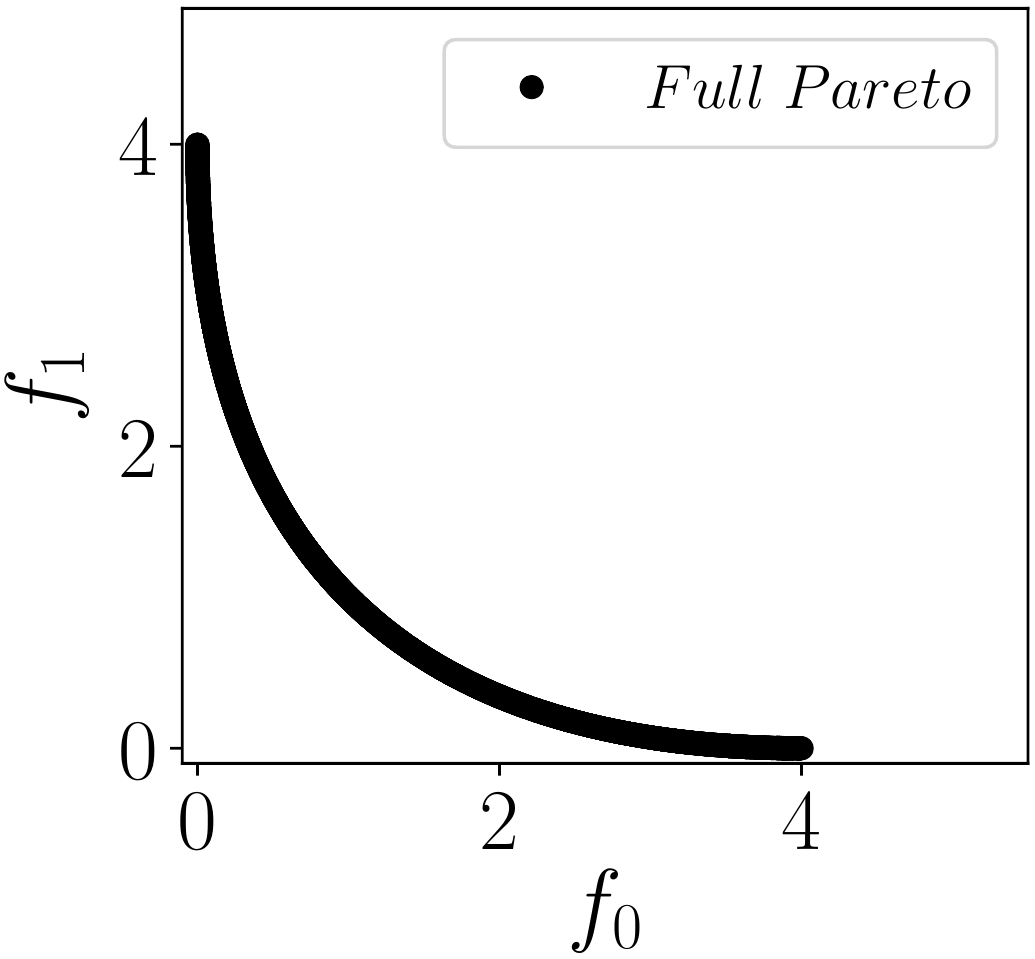}
    \caption{Full Pareto front}
    \label{fig:mainS}
\end{subfigure}
\begin{subfigure}[t]{0.315\linewidth}
    \centering
    \includegraphics[width=1\linewidth]{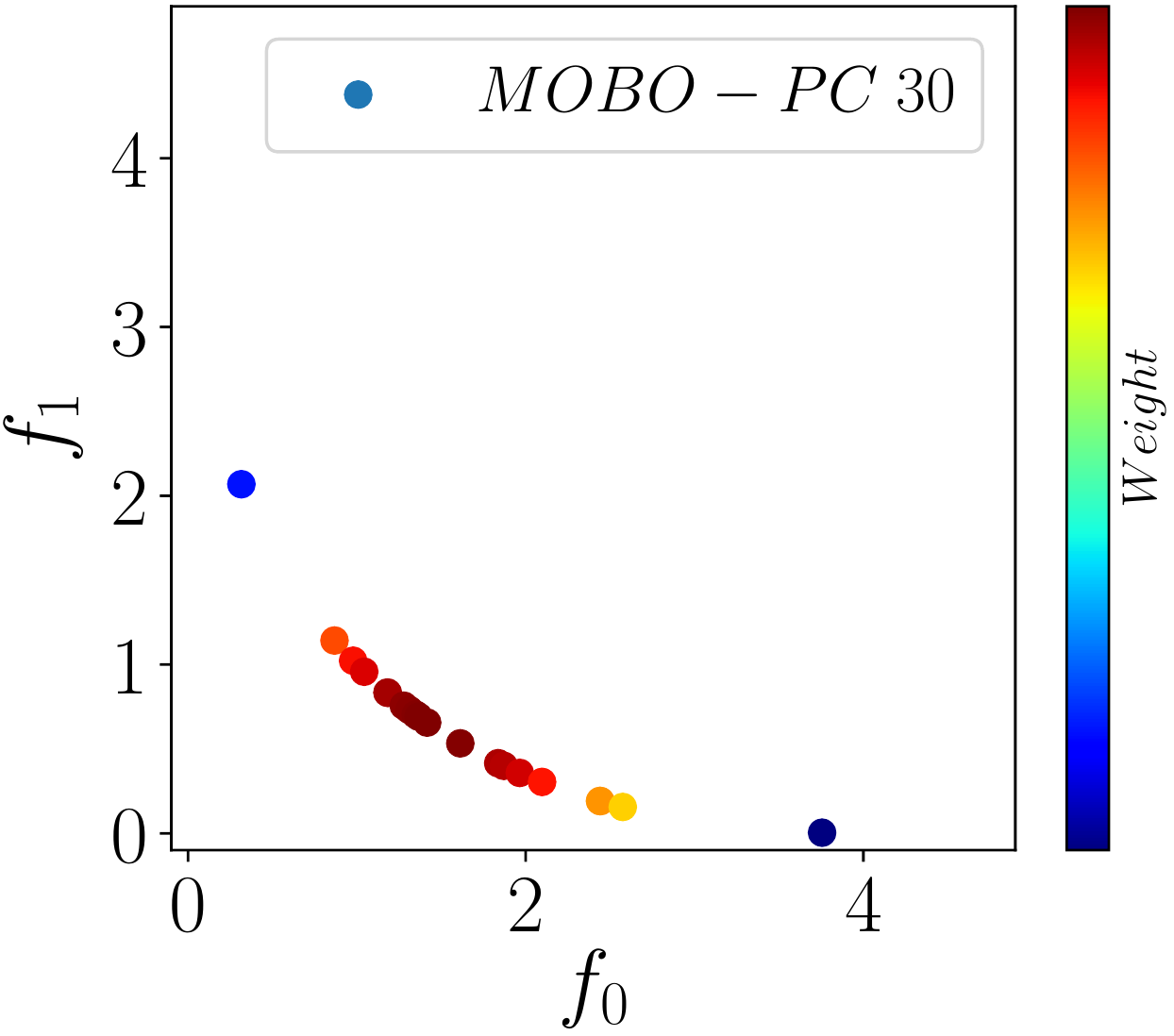}
    \caption{Case $1$, $s_{0} \approx s_{1}$}
    \label{fig:mainS1}       
\end{subfigure}
\begin{subfigure}[t]{0.315\linewidth}
    \centering
    \includegraphics[width=1\linewidth]{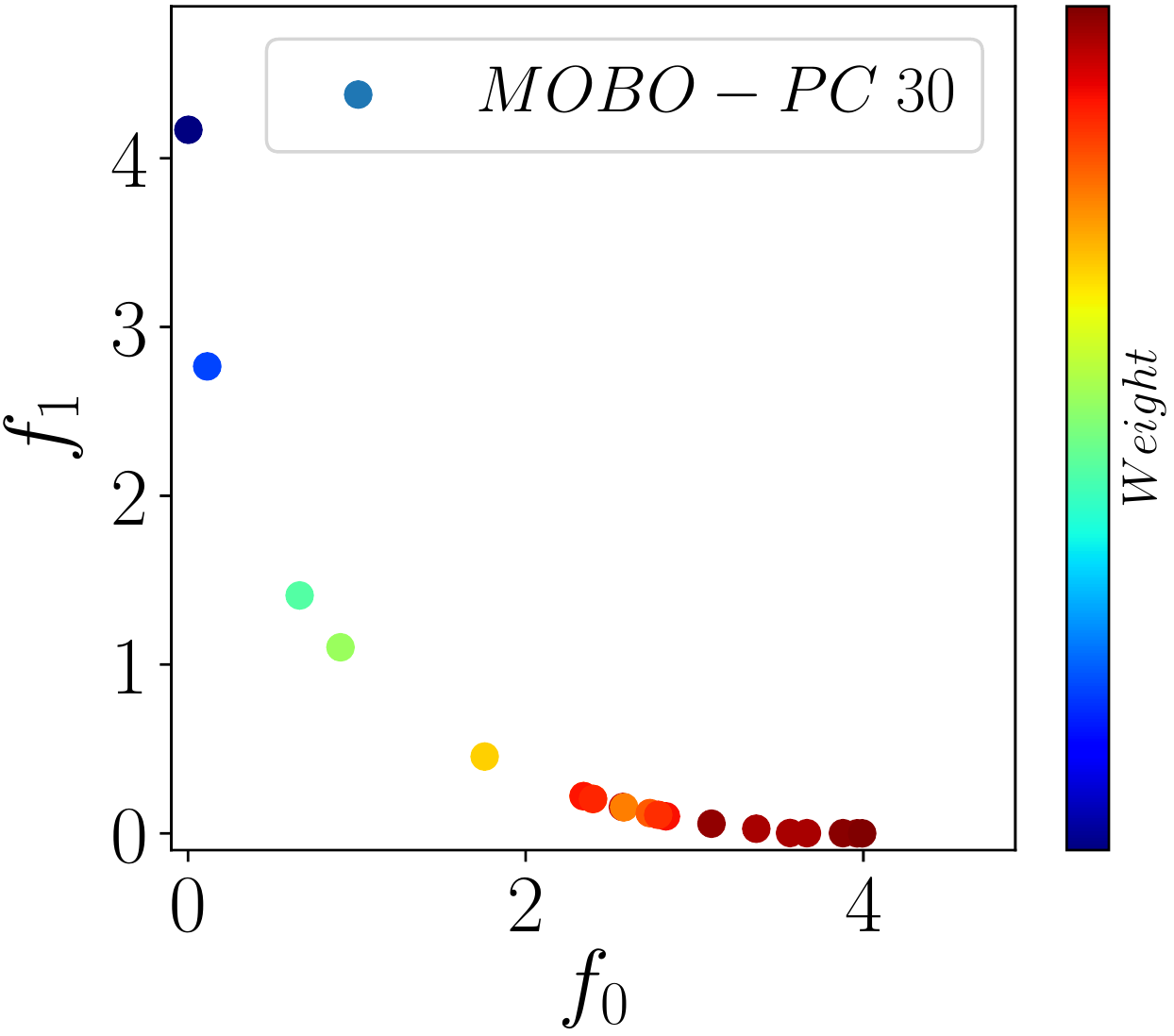}
    \caption{Case $2$, $s_{0} < s_{1}$}
    \label{fig:mainS2}      
\end{subfigure}
\begin{subfigure}[t]{0.315\linewidth}
    \centering
    \includegraphics[width=1\linewidth]{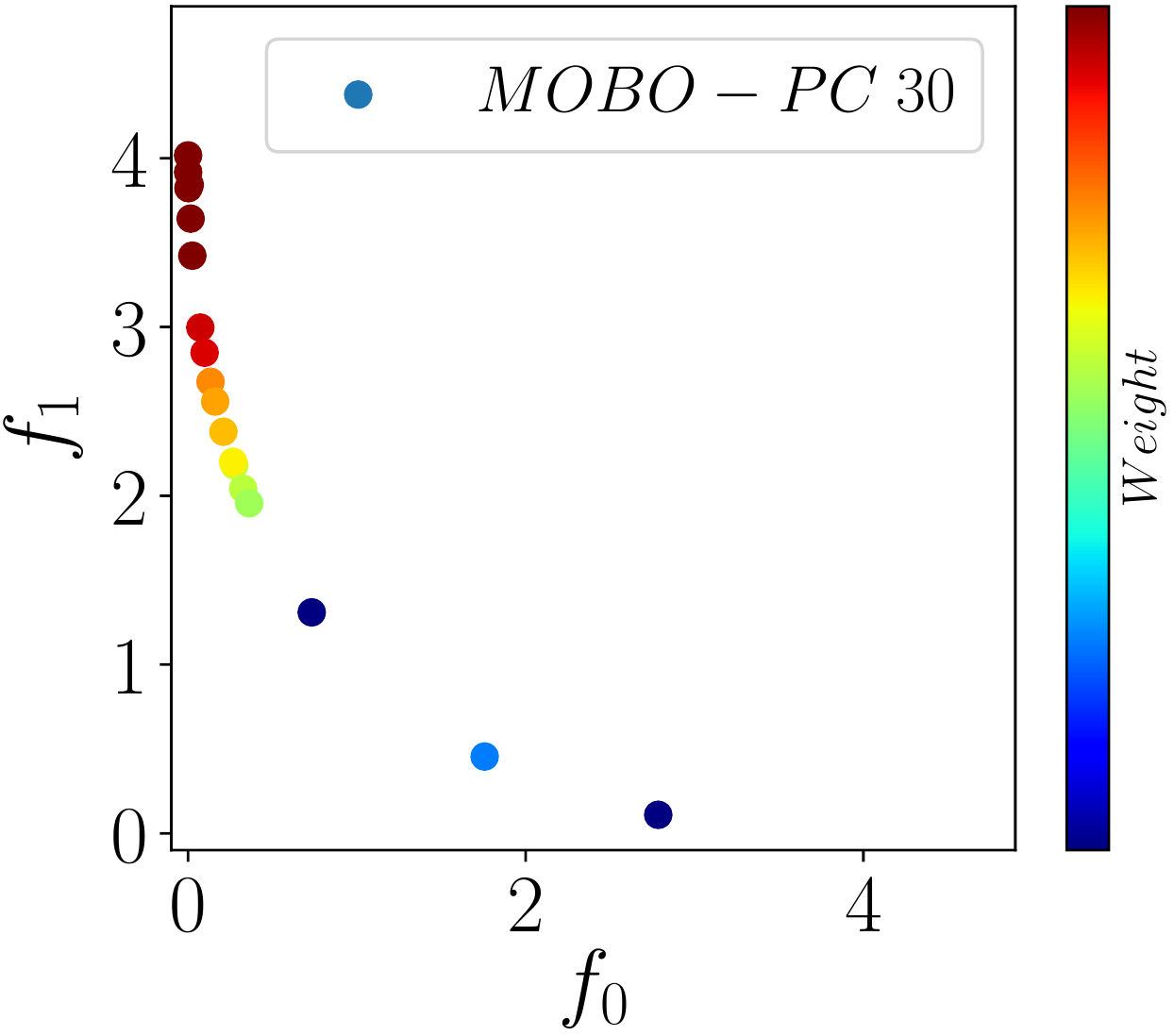}
    \caption{Case $3$, $s_{0} > s_{1}$}
    \label{fig:mainS3}      
\end{subfigure}
\begin{subfigure}[t]{0.281\linewidth}
    \centering
    \includegraphics[width=1\linewidth]{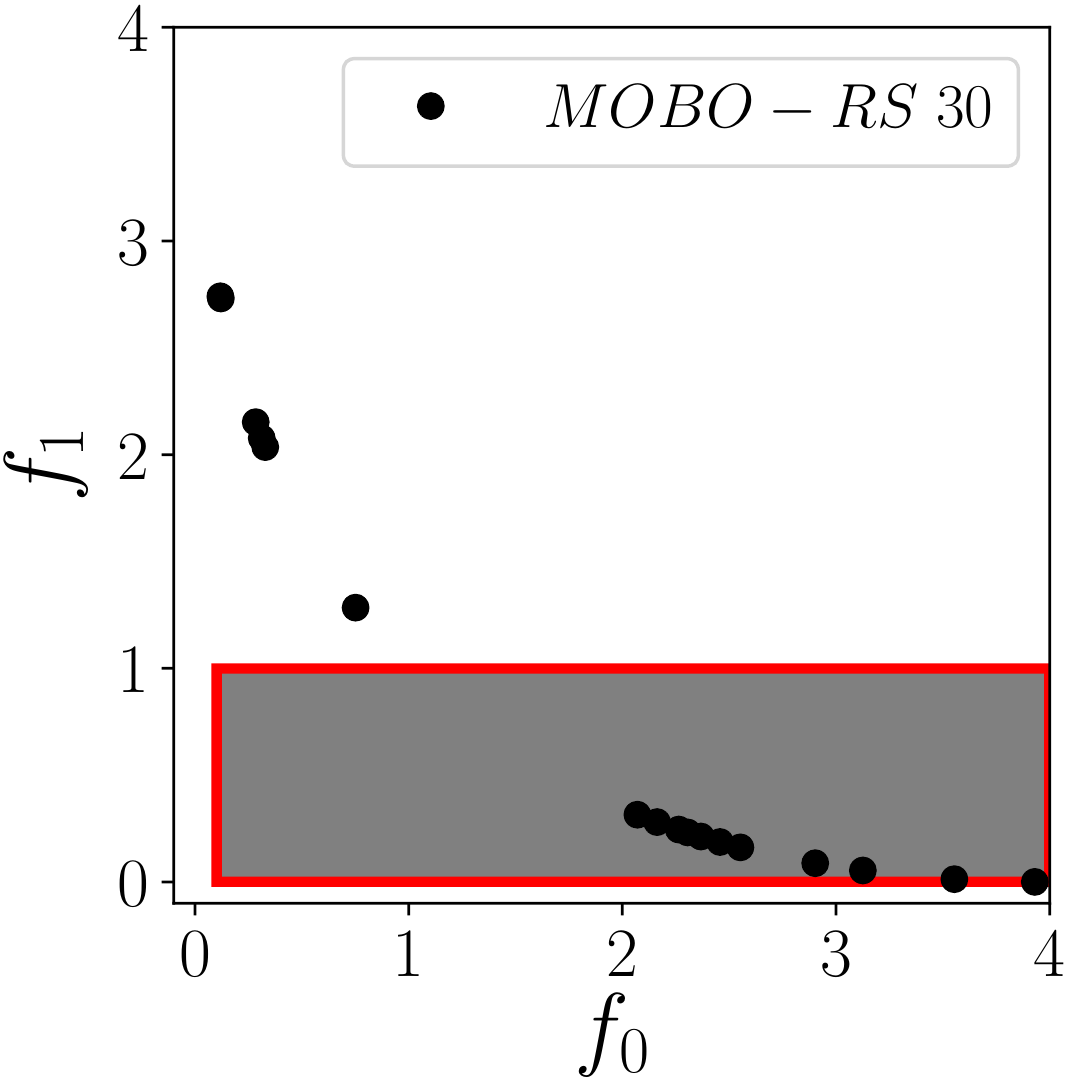}
\caption{}
    \label{fig:mainK11}      
\end{subfigure}
\begin{subfigure}[t]{0.281\linewidth}
    \centering
    \includegraphics[width=1\linewidth]{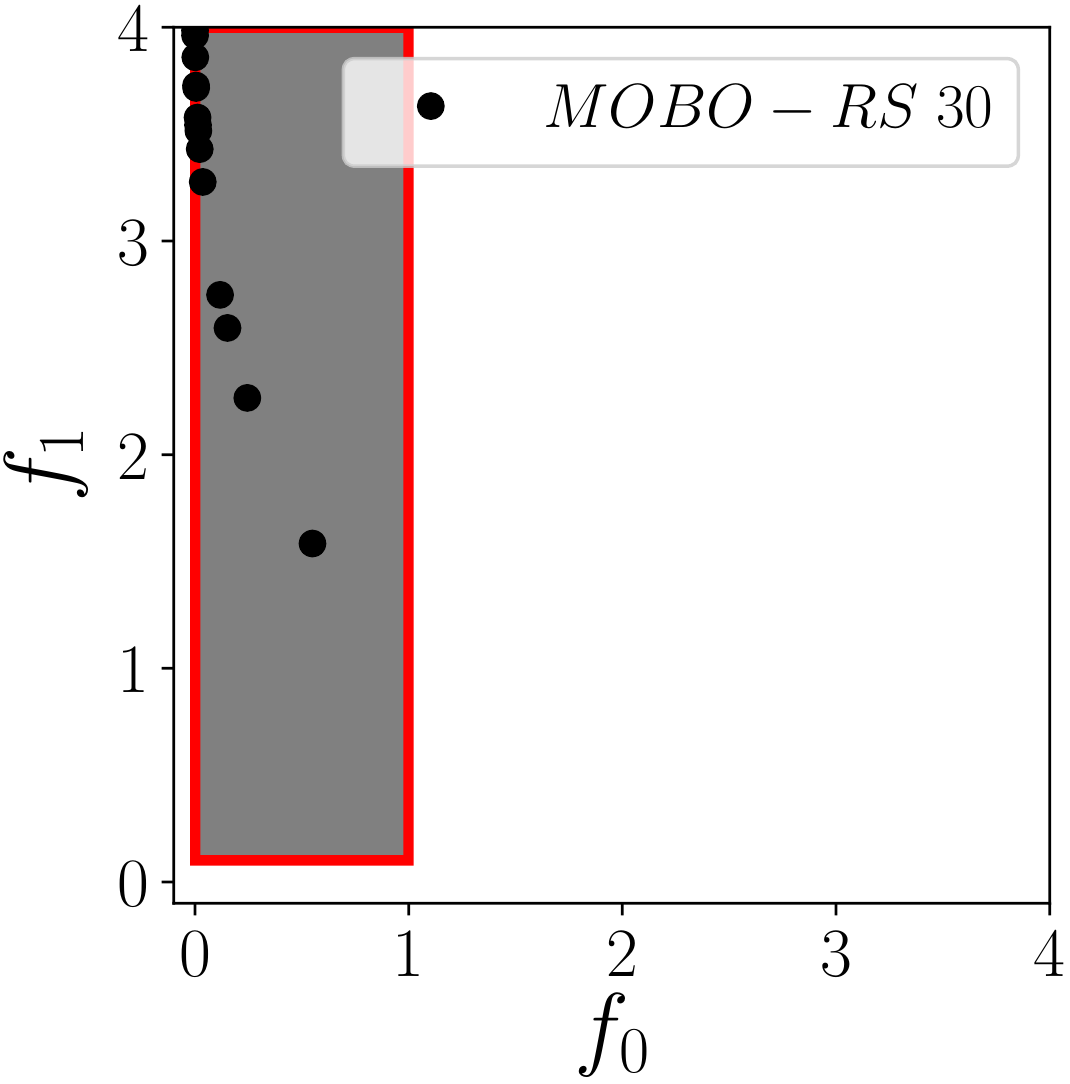}

\caption{}
    \label{fig:mainK12}      
\end{subfigure}
\caption{Finding Pareto front which comply with the preference-order constraint. Figure \ref{fig:mainS} shows the full Pareto front solution (with no preferences). Figure \ref{fig:mainS1} illustrates the Pareto front by assuming stability of first objective $f_0$ is similar to second objective $f_1$. In Figure \ref{fig:mainS2}, stability of $f_1$ is preferred over $f_0$. Figure \ref{fig:mainS3} shows more stable results for $f_0$ than $f_1$ ($s_0>s_1$). Figure \ref{fig:mainK11} and \ref{fig:mainK12} shows the results obtained by MOBO-RS and the corresponding bounding boxes. The gradient color of the Pareto front points in Figure \ref{fig:mainS1}-\ref{fig:mainS3} indicates their degree of compliance with the constraints.}
\end{figure}

\subsection{Synthetic Functions}
We begin with a comparison on minimising synthetic function Schaffer function N. $1$ with $2$ conflicting objectives $f_0$, $f_1$ and $1$-dimensional input. (see \cite{schaffer1984some}). 
Figure \ref{fig:mainS} shows the ground-truth Pareto front for this function. To illustrate the behavior of our 
method, we impose distinct preferences.  Three test cases are designed to illustrate the  effects of imposing preference-order constraints on the objective functions for stability. 
Case ($1$): $s_{0} \approx s_{1}$, Case ($2$): $s_{0} < s_{1}$ and Case ($3$): $s_{0} > s_{1}$. For our method it is only required to define the preference-order constraints, however for MOBO-RS, additional information as a bounding box is obligatory. Figure \ref{fig:mainS1} (case $1$), shows the results of preference-order constraints $ \begin{array}{l}  \infsetstyle{S}_{\ordsetstyle{I}} \triangleq \left\{ \left. {\bf s} \in \posquad \backslash \left\{ {\bf 0} \right\} \right| s_{0} \approx s_{1} \right\}\end{array}$ for our proposed method,
 where $s_{0}$  represents the importance of stability in minimising  $f_0$ and $s_{1}$ is the importance of stability in minimising $f_1$. Due to same importance of both objectives, a balanced optimisation is expected. Higher weights are obtained for the Pareto front points in the middle region with highest stability for both objectives. Figure \ref{fig:mainS2} (case $2$) is based on the preference-order of $s_{0} < s_{1}$ that implies the importance of stability in $f_1$ is more than $f_0$. The results show more stable Pareto points for $f_1$ than $f_0$. Figure \ref{fig:mainS3} (case $3$) shows the results of $s_{0} > s_{1}$ preference-order constraint. As expected, we see more number of stable Pareto points for the important objective (i.e. $f_0$ in this case). We defined two bounding boxes for MOBO-RS approach which can represent the preference-order constraints in our approach (Figure \ref{fig:mainK11} and \ref{fig:mainK12}). There are infinite possible bounding boxes can serve as constraints on objectives in such problems, consequently, the instability of results is expected across the various definitions of bounding boxes. We believe our method can obtain more stable Pareto front solutions especially when prior information is sparse. Also, having extra information as the weight (importance) of the Pareto front points is another advantage.

\begin{figure*}[t]
\centering
\label{fig:Mould}
\begin{subfigure}[t]{0.325\linewidth}
    \centering
    \includegraphics[width=1\linewidth]{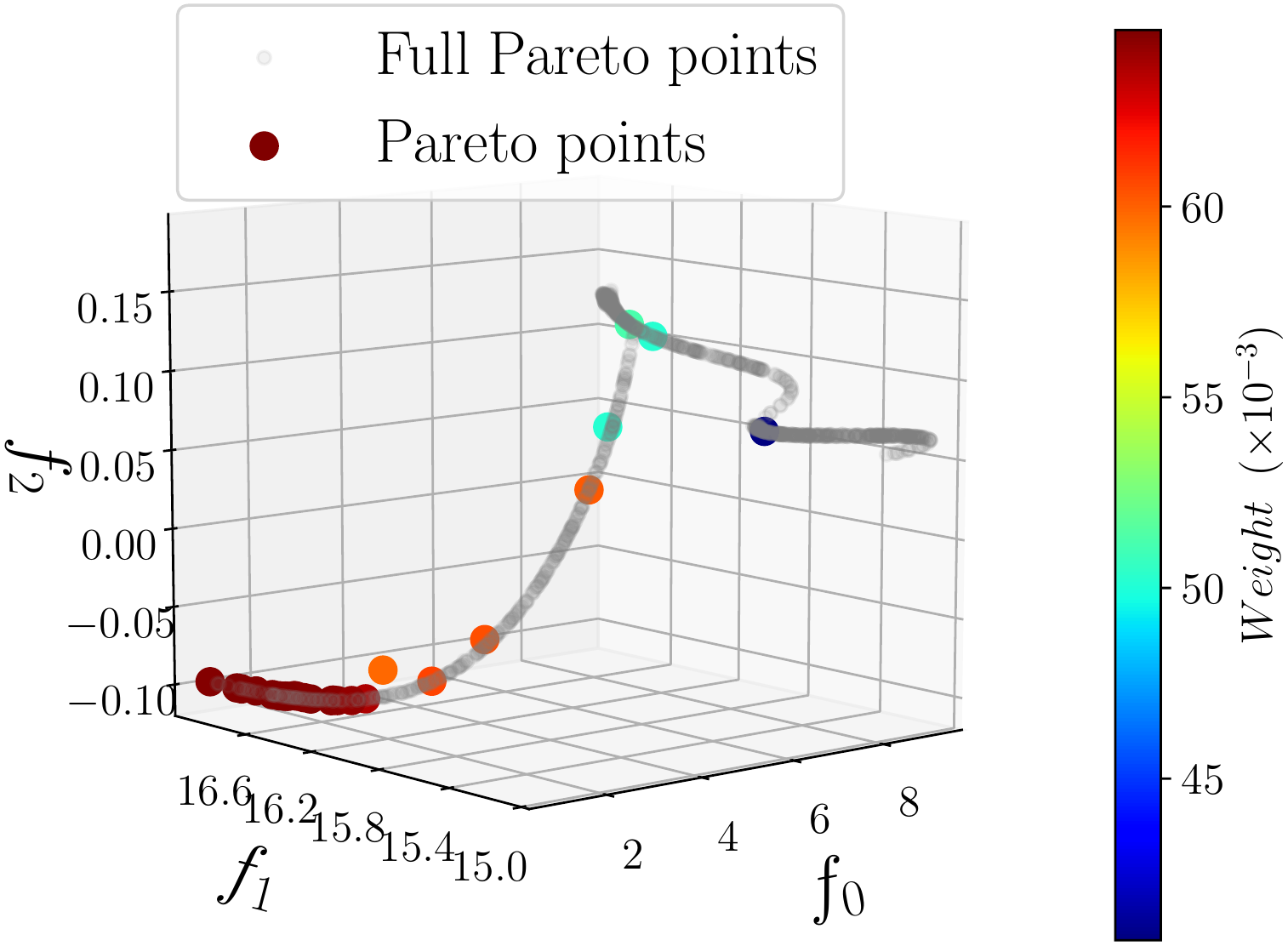}
    \caption{Obtained Pareto points}
    \label{fig:mainS_3d}
\end{subfigure}
\begin{subfigure}[t]{0.325\linewidth}
    \centering
    \includegraphics[width=1\linewidth]{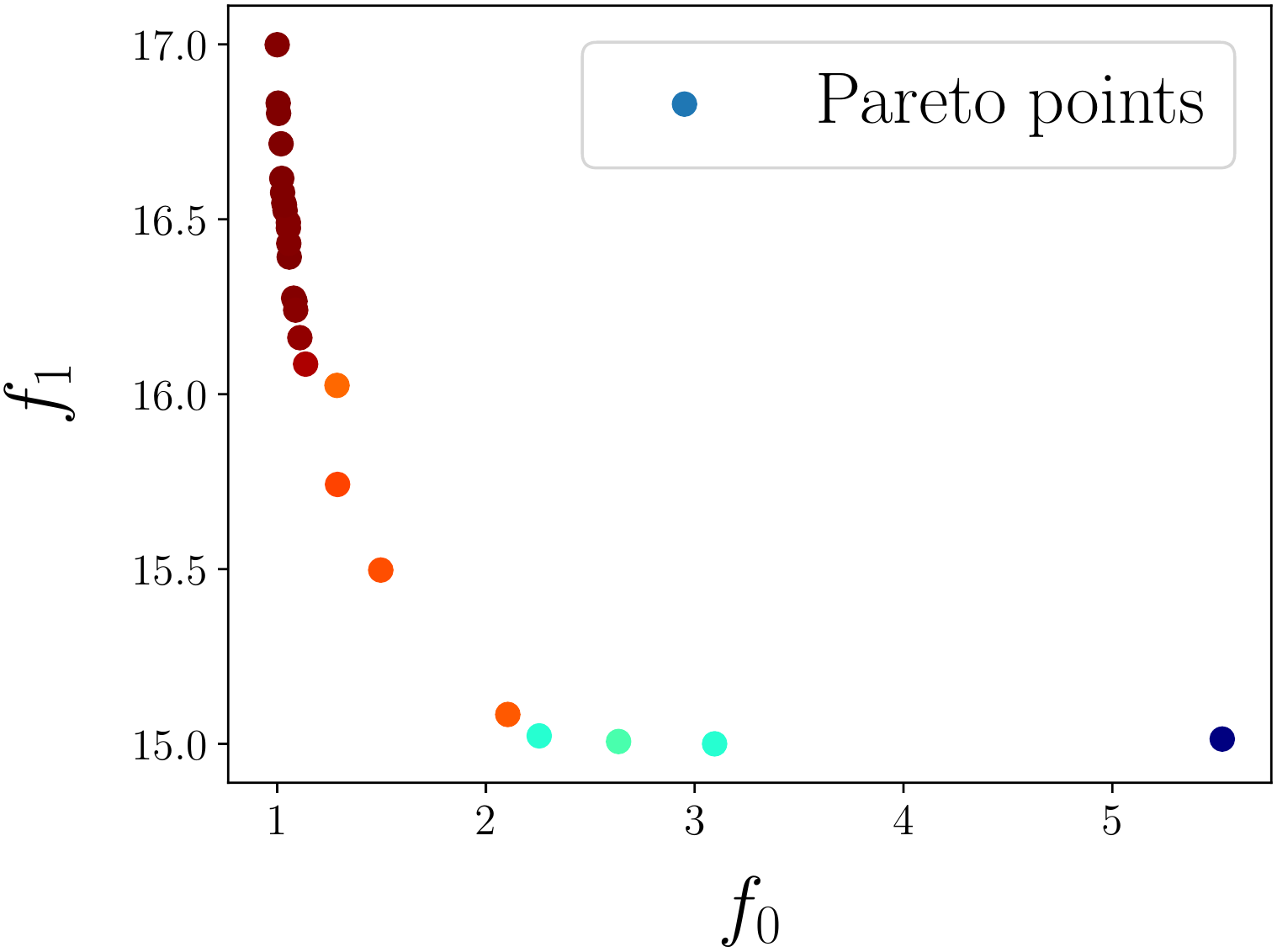}
    \caption{Projection of Pareto points in $f_0$ and $f_1$ space}
    \label{fig:mainS1_3d}       
\end{subfigure}
\begin{subfigure}[t]{0.325\linewidth}
    \centering
    \includegraphics[width=1\linewidth]{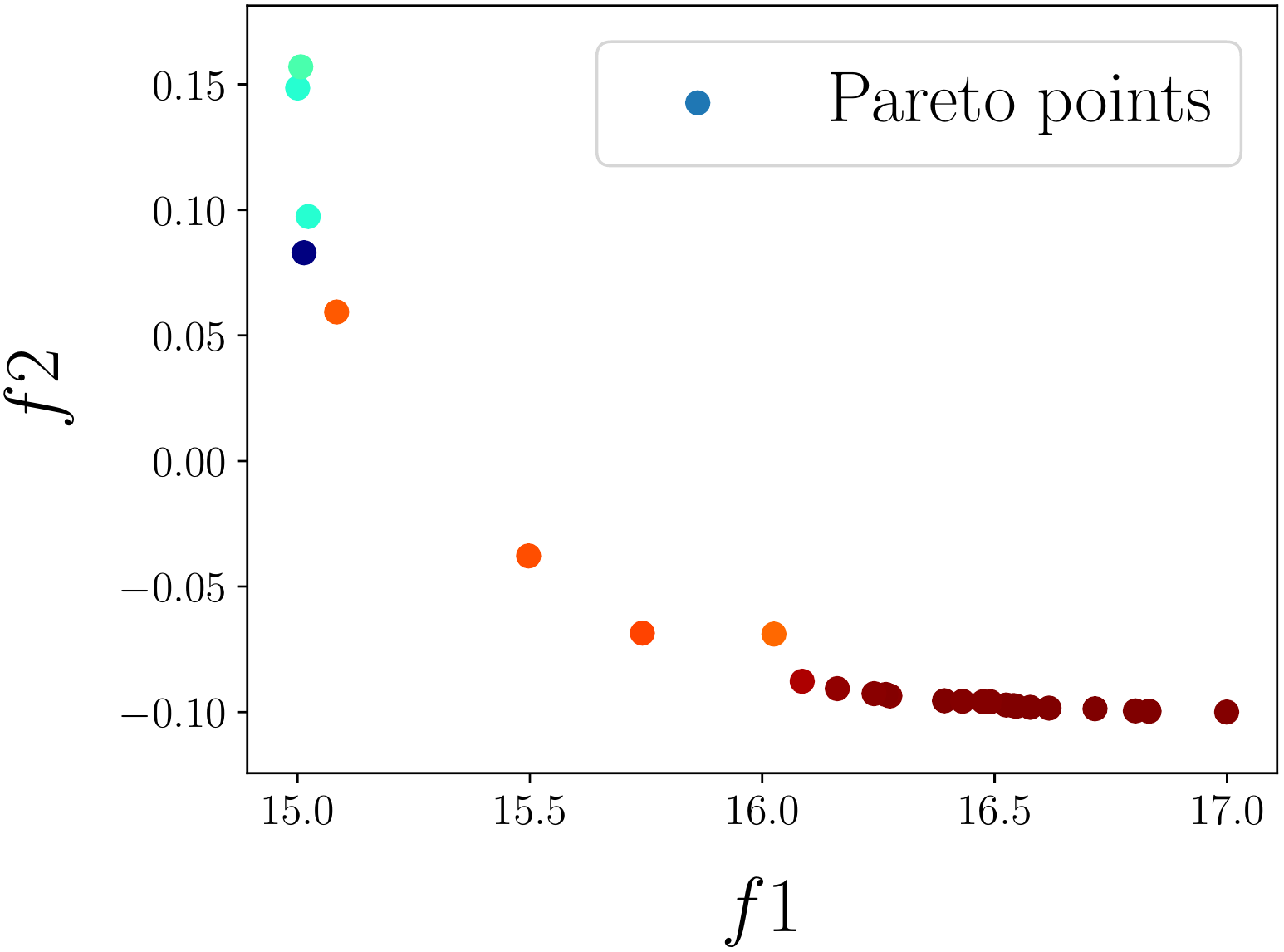}
    \caption{Projection of Pareto points in $f_1$ and $f_2$ space}
    \label{fig:mainS2_3d}      
\end{subfigure}
\caption{Finding Pareto front points with partial constraints as specified by $s_0>s_1$ and $s_2>s_1$. Figure \ref{fig:mainS_3d} shows the $3$D plot of the obtained Pareto front points satisfying preference-order constraints with the color indicating the degree of compliance. Figure \ref{fig:mainS1_3d} illustrates the projection of Pareto optimal points on $f_0\times f_1$ sub-space, and figure \ref{fig:mainS2_3d} shows the projection on $f_1\times f_2$ sub-space.}
\label{fig:3dMain}
\end{figure*}
\begin{figure*}[t]
         \centering\noindent
\includegraphics[scale=0.45]{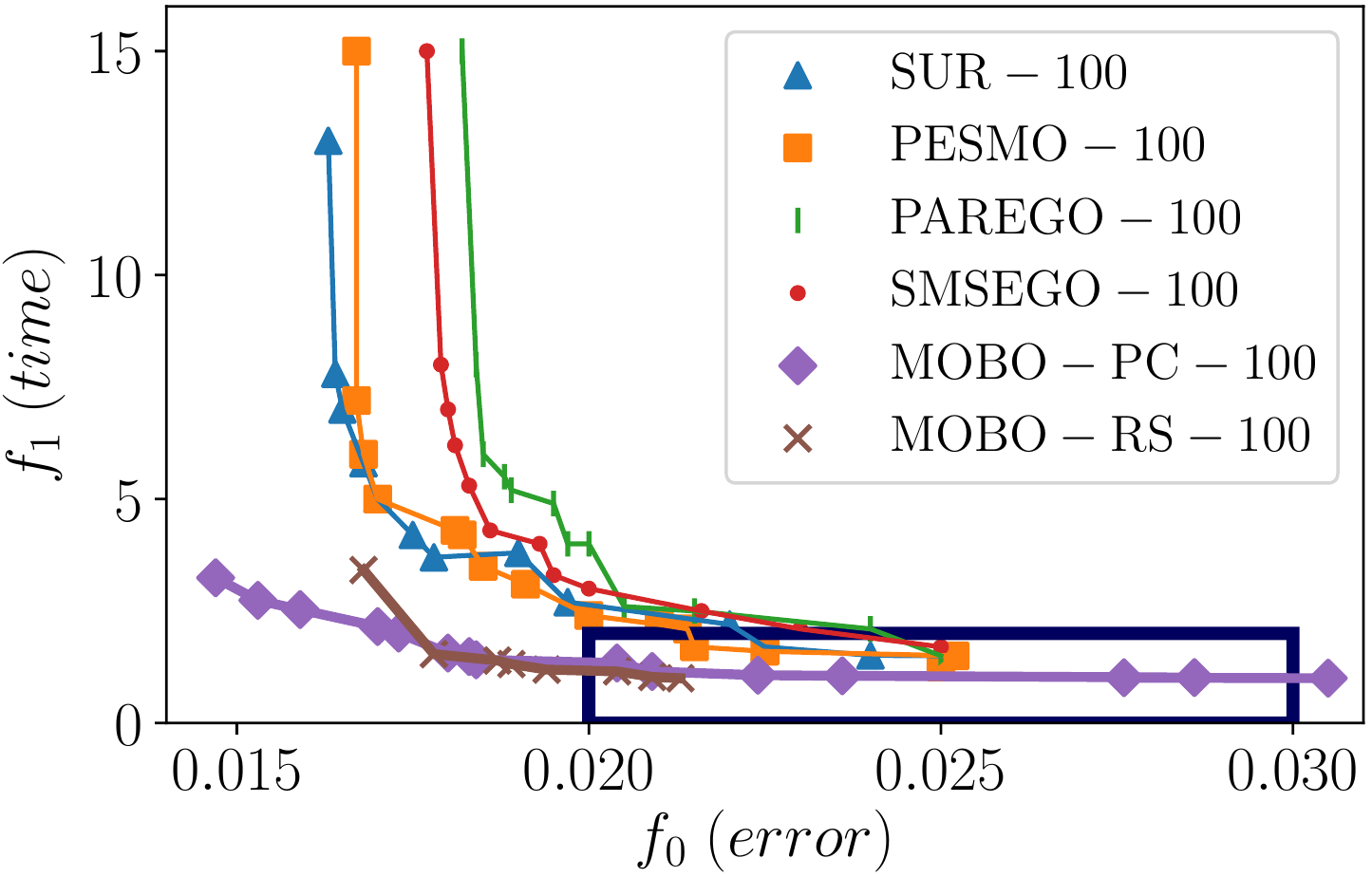}
\includegraphics[scale=0.45]{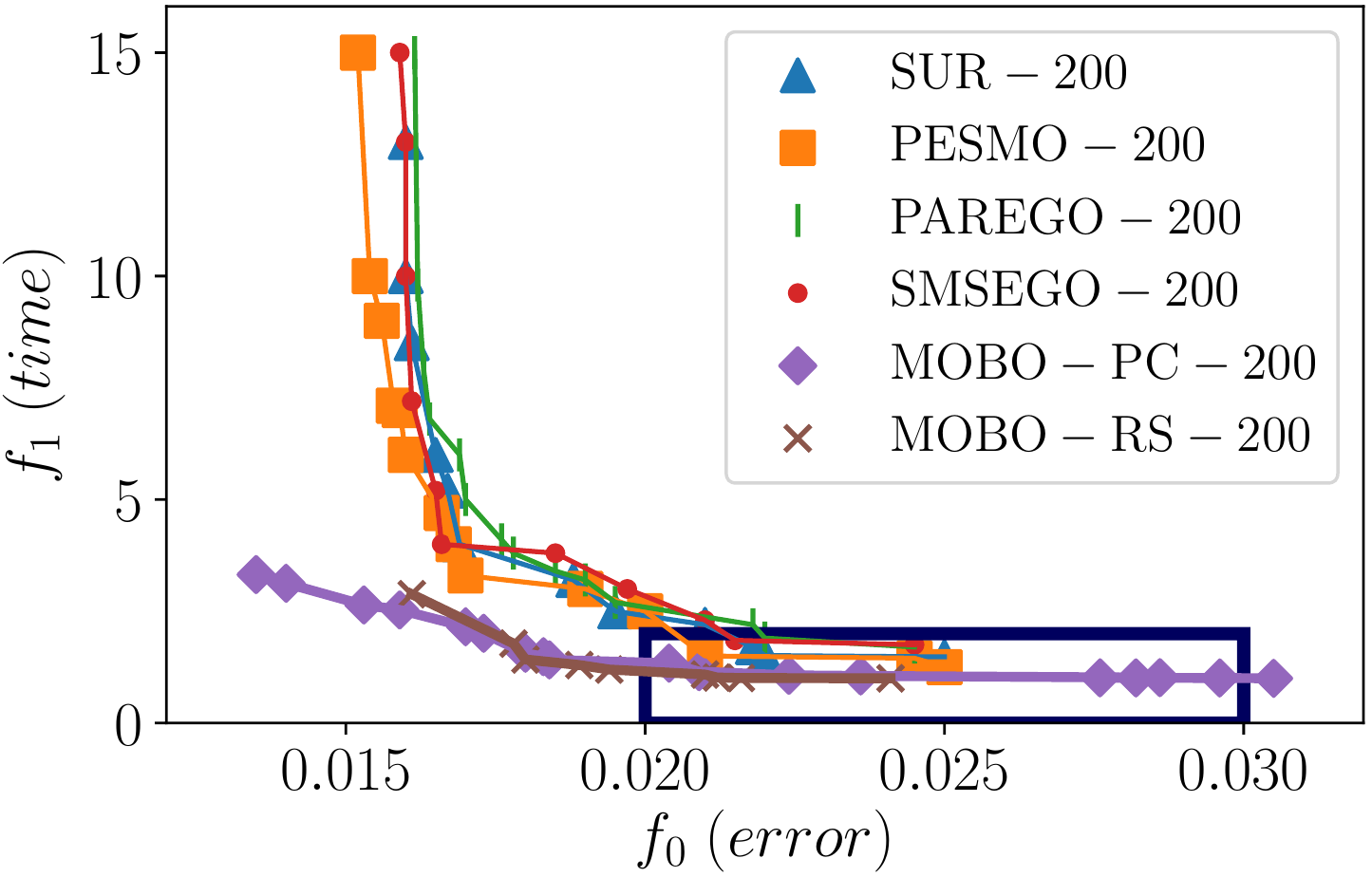}
         \captionof{figure}{Average Pareto fronts obtained by proposed method in comparison to other methods. This experiment defines $s_{1} > s_{0}$ i.e. stability of run time is more important than the error. For MOBO-RS, $[[0.02,0],[0.03,2]]$ is an additional information used as bounding box. The other methods do not incorporate preferences. The results are shown for $100$ evaluations of the objectives (left) and $200$ evaluations of the objectives (right).\vspace{-0.7cm}}
\label{fig:E4}
\end{figure*}
\par
Figure \ref{fig:3dMain} illustrates a special test case in which $s_0 > s_1$ and $s_2 > s_1$, yet no preferences specified over $f_2$ and $f_0$ while minimising Viennet function. The proposed complex preference-order constraint does not form a proper cone as elaborated in Theorem \ref{th:th_geomofS}. However, $s_0 > s_1$ independently constructs a proper cone, likewise for $s_2 > s_1$. Figure \ref{fig:mainS_3d} shows the results of processing these two independent constraints separately, merging their results and finding the Pareto front. Figure \ref{fig:mainS1_3d} 
implies more stable solutions for $f_0$ comparing to $f_1$. Figure \ref{fig:mainS2_3d} shows the Pareto front points comply with $s_2 > s_1$.
\subsection{Finding a Fast and Accurate Neural Network}
\vspace{-0.1cm}
Next, we train a neural network with two objectives of minimising both prediction error and prediction time, as per \cite{hernandez2016predictive}. These are conflicting objectives because reducing the prediction error generally involves larger networks and consequently longer testing time. We are using MNIST dataset and the tuning parameters include number of hidden layers ($x_1 \in [1,3]$), the number of hidden units per layer ($x_2 \in [50,300]$), the learning rate ($x_3 \in (0,0.2]$), amount of dropout ($x_4 \in [0.4,0.8]$), and the level of $l_1\ (x_5 \in (0,0.1])$ and $l_2\ (x_6 \in (0,0.1])$ regularization. For this problem we assume stability of $f_1 (time)$ in minimising procedure is more important than the $f_0(error)$. For MOBO-RS method, we selected $[[0.02,0],[0.03,2]]$ bounding box to represent an accurate prior knowledge (see Figure \ref{fig:E4}). The results were averaged over $5$ independent runs. Figure \ref{fig:E4} illustrates that one can simply ask for more stable solutions with respect to test time (without any prior knowledge) of a neural network while optimising the hyperparameters.  As all the solutions found with MOBO-PC are in range of $(0,5)$ test time. In addition, it seems the proposed method finds more number of Pareto front solutions in comparison with MOBO-RS. \vspace{-0.5cm}

\section{Conclusion}
\vspace{-0.35cm}
In this paper we proposed a novel multi-objective Bayesian optimisation algorithm with preferences over objectives. We define objective preferences in terms of stability and formulate a common framework to focus on the sections of the Pareto front where preferred objectives are more stable, as is required. 
We evaluate our method on both synthetic and real-world problems and show that the obtained Pareto fronts comply with the preference-order constraints.
\vspace{-0.4cm}
\section*{Acknowledgments}
\vspace{-0.3cm}
This research was partially funded by Australian Government through the Australian Research Council (ARC). Prof Venkatesh is the recipient of an ARC Australian Laureate Fellowship $\mathrm{(FL170100006)}$.

\bibliographystyle{plain}
\bibliography{neurips_2019}
\clearpage
\begin{center}
\centering
\Large{\bf Supplementary Materials}
\end{center}
\section{Method of Evaluation}
For a more precise evaluation of our proposed method, we can define a measurement by checking how many of the Pareto front solutions satisfy the preference-order constraints. Based on Algorithm 3, we can calculate the percentage of solutions that satisfy the preference-order constraints by using the gradients of the actual synthetic functions at iteration $t$. In real-world problems, we may use the gradients of the trained Gaussian Process to evaluate the compliance of Pareto front with preference-order constraints. 
\par
\begin{figure}
\centering
\label{fig:Mould}
\begin{subfigure}[t]{0.4\linewidth}
    \centering
    \includegraphics[width=1\linewidth]{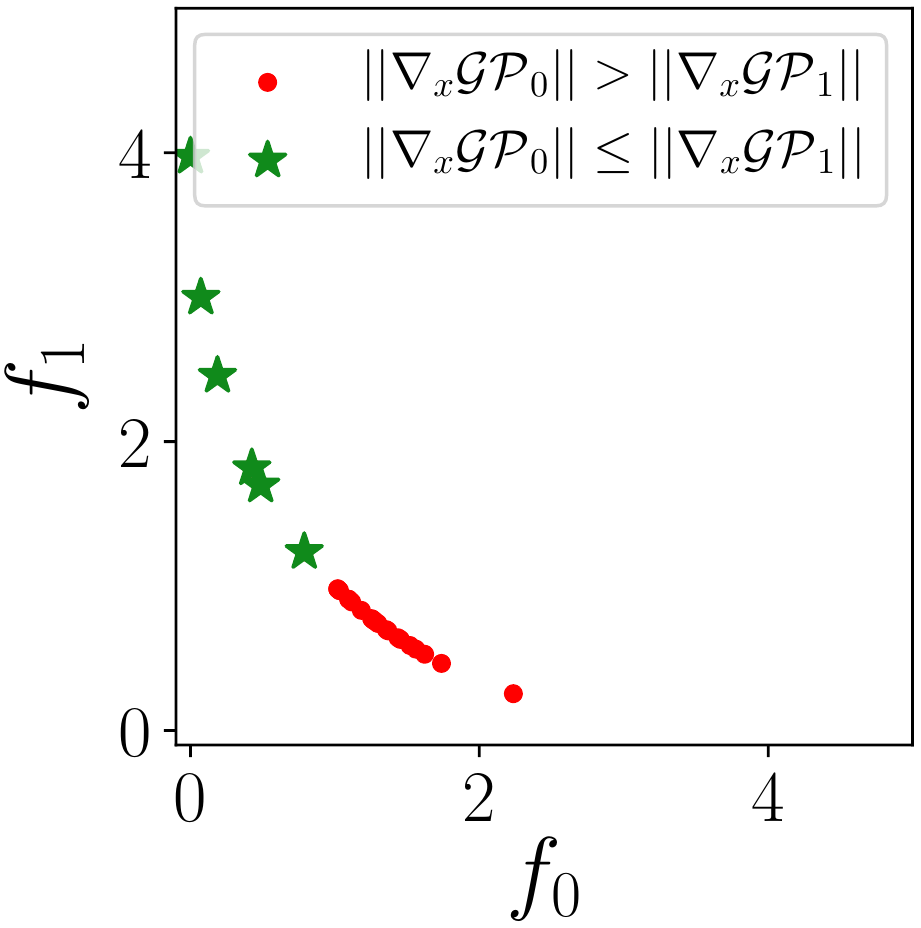}
    \caption{MOBO-PC}
\end{subfigure}
\begin{subfigure}[t]{0.4\linewidth}
    \centering
    \includegraphics[width=1\linewidth]{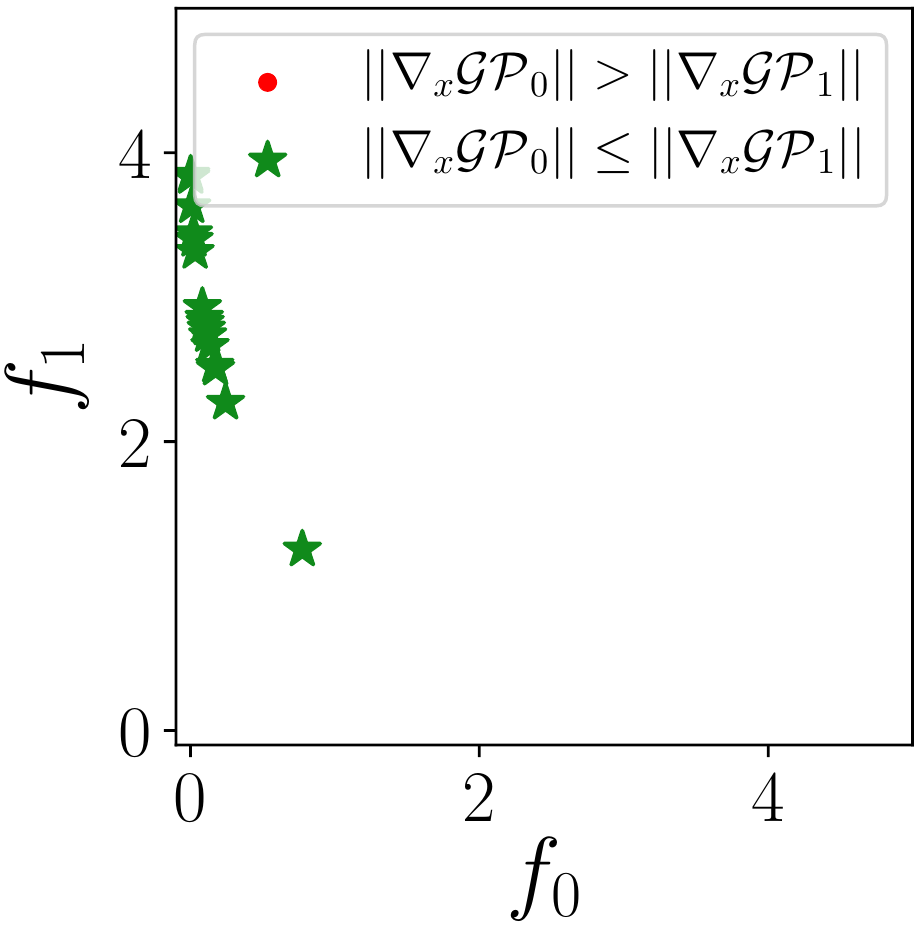}
    \caption{Naive Approach}
\end{subfigure}

\caption{(a) Illustrates a naive approach of post-processing in which almost 50\% of the solutions are not complying with preference-order constraints and must be disregarded. (b) Shows an obtained Pareto front by MOBO-PC. All of the obtained solutions are complying with the preference-order constraints based on the  derivative of the trained Gaussian Processes.}
\label{fig:Patch}      
\end{figure}

It is noteworthy to mention that our function evaluations are expensive, and hence, throwing away evaluations during post-processing is undesirable. Our approach, in contrast, samples such that most of the function evaluations would have desirable characteristics, and hence, would be efficient. Considering Figure \ref{fig:Patch}, given a preference-order constraint as ``stability of $f_0$ being more important than $f_1$'' in Schaffer function N. 1, i.e. $||\frac{\partial f_0}{\partial x}|| \leq ||\frac{\partial f_1}{\partial x}||$, Figure \ref{fig:Patch} (a) illustrates the Pareto front obtained by a { plain multi-objective optimisation} (with no constraints). After the Pareto solutions are found (in $20$ iterations), using the derivatives of the trained Gaussian Processes (actual objective functions are black-box), we can post process the obtained Pareto front based on the stability of solutions. Figure \ref{fig:Patch} (a) shows that only $\frac{6}{18}$ of these solutions have actually met the preference-order constraints. Whereas Figure \ref{fig:Patch} (b) shows that $\frac{16}{16}$ of the obtained Pareto front solutions by MOBO-PC (in the same 20 iterations) have met the preference-order constraints. 
\par
In general, our experimental results show 98.8\% of solutions found for Schaffer function N.
1 after 20 iterations comply with constraints. As for Poloni's two objective function, 86.3\% of the solutions follow the constraints after 200 iterations and finally for Viennet 3D function, 
this number is 82.5\%. 
\par
Given that the prior knowledge is not provided in \cite{Kand2018}, the obtained results for their method with same experimental
design and the same number of iterations are 47.2\% for Schaffer function N. 1, 29.6\% for Poloni’s two objective function
and 19.3\% for Viennet 3D function respectively. This gap explains the importance of the prior knowledge about hyperboxes for their method. The reported numbers are averaged over 10 independent runs. Table \ref{tab:Patch} summarises the obtained results.
\ \\ \\
\begin{minipage}{\linewidth}
\centering
\captionof{table}{Percentage of the Pareto front solutions complying with preference-order constraints in different synthetic functions.} \label{tab:Patch} 
\begin{tabularx}{\linewidth}{@{} C{1in} C{.65in} *3X @{}}\toprule[2.0pt]
{ } & \bf Schaffer function N. 1 & \bf Poloni & \bf Viennet 3D \\\midrule
\bf MOBO-PC & $98.8\%$ & $86.3\%$ & $82.5\%$ \\\midrule
\bf MOBO-RS & $47.2\%$ & $29.6\%$ & $19.3\%$ \\
\bottomrule[2.0pt]
\end {tabularx}\par
\bigskip
\end{minipage}

\section{Proofs}
\begin{th_geomofS} 
 Let $\ordsetstyle{I} = (0, 1, \ldots,Q| Q \in \infsetstyle{Z}_m \textbackslash \{0\})$ be an (ordered) 
 preference tuple.  Define $\infsetstyle{S}_{\ordsetstyle{I}}$ as per 
 definition $1$.  Then $\infsetstyle{S}_{\ordsetstyle{I}}$ 
 is a polyhedral (finitely-generated) proper cone (excluding the origin) that 
 may be represented using either a polyhedral representation:
 \begin{equation}
  \begin{array}{l}
   \infsetstyle{S}_{\ordsetstyle{I}} = \left\{ \left. {\bf s} \in \infsetstyle{R}^m \right| {\bf a}_{(i)}^{\tsp} {\bf s} \geq 0 \forall i \in \infsetstyle{Z}_m \right\} \backslash \left\{ {\bf 0} \right\} \\
  \end{array}
  \label{eq:defSpoly}
 \end{equation} 
 or a generative representation:
 \begin{equation}
  \begin{array}{rl}
   \infsetstyle{S}_{\ordsetstyle{I}} &\!\!\!\!= \Big\{ \mathop{\sum}\limits_{i \in \infsetstyle{Z}_m} c_i \tilde{\bf a}_{(i)} \left|\; {\bf c} \in \posquad \right. \Big\} \backslash \left\{ {\bf 0} \right\} \\
  \end{array}
  \label{eq:defSfingen}
 \end{equation} 
 where $\forall i \in \infsetstyle{Z}_m$:
 \[
  \begin{array}{l}
  \begin{array}{rl}
   {\bf a}_{(i)} &\!\!\!\!\!= \left\{ \!\!\begin{array}{ll} 
    \frac{1}{\sqrt{2}} \left( {{\bf e}_{i} - {\bf e}_{i+1}} \right) & \!\!\mbox{if } i \in \infsetstyle{Z}_Q \\ 
    {\bf e}_{i}                                                             & \!\!\mbox{otherwise} \\ 
   \end{array} \!\!\right. \\
   \tilde{\bf a}_{(i)} &\!\!\!\!\!= \left\{ \!\!\begin{array}{ll} 
    \frac{1}{\sqrt{i+1}} \mathop{\sum}\limits_{ l  \in \infsetstyle{Z}_{i+1}} {\bf e}_{l} & \!\!\mbox{if } i \in \infsetstyle{Z}_{Q+1} \\ 
    {\bf e}_{i}                                                                           & \!\!\mbox{otherwise} \\ 
   \end{array} \!\!\right. \\
  \end{array}
  \end{array}
 \]
 and ${\bf e}_{0}, {\bf e}_{1}, \ldots, {\bf e}_{m-1}$ are the Euclidean 
 basis of $\infsetstyle{R}^m$.
 \label{th:th_geomofS}
\end{th_geomofS}

\begin{proof}
The polyhedral representation follows directly from consideration of the 
constraints $s_i \geq 0$, from which we derive the constraints ${\bf 
a}_{(i)}^{\tsp} {\bf s} = s_{i} \geq 0$ $\forall i \notin \infsetstyle{Z}_Q$; 
and the constraints $s_{k} \geq s_{{k+1}}$ $\forall k \in 
\infsetstyle{Z}_Q$, from which we derive the constraints ${\bf 
a}_{(k)}^{\tsp} {\bf s} = s_{k} - s_{{k+1}} \geq 0$ $\forall k = 0, 1, 
\ldots, Q-1$.  Moreover $\infsetstyle{S}_{\ordsetstyle{I}} \cup \{ {\bf 0} \} 
\subset \posquad$ is constructed by restricting $\posquad$ using half-space 
constraints, so $\infsetstyle{S}_{\ordsetstyle{I}} \cup \{ {\bf 0} \}$ is a 
proper cone.

The generative representation follows from the fact that a proper conic 
polyhedra is positively spanned by it's extreme directions ($\tilde{\bf a}_{(i)}$) - i.e. the 
intersections of the hyperplanes ${\bf a}_{(i)}^\tsp {\bf s} = 0$.  So $\forall i \in \infsetstyle{Z}_m$, 
$\tilde{\bf a}_{(i)}$ must satisfy:
\begin{equation}
 \begin{array}{l}
  \tilde{\bf a}_{(i)}^{\tsp} {\bf a}_{(j)} = 0 \; \forall j \ne i
 \end{array}
 \label{eq:tildeacond}
\end{equation}
There are six cases that are possible combinations of $i$ and $j$. We show how  (\ref{eq:tildeacond}) holds in all the conditions.
\begin{enumerate}
\item $i \in \{ 0, 1, \ldots, Q-1 \}$ and $j \in \{ Q, Q+1, \ldots, m-1 \}$: considering theorem \ref{th:th_geomofS}, $\tilde{\bf a}_{(i)}^{\tsp} {\bf a}_{(j)} = \tilde{\bf a}_{(i)}^{\tsp}{\bf e}_{j} = 0$. Therefore $\tilde{\bf a}_{(i)j} = 0$. Which implies $\tilde{\bf a}_{(i)Q} = \tilde{\bf a}_{(i)Q+1} = \ldots = \tilde{\bf a}_{(i)m-1} = 0$. 
\item $i \in \{ 0, 1, \ldots, Q-1 \}$ and $j \in \{ 0, 1, \ldots, Q-1 \}\textbackslash \{i\}$: based on theorem \ref{th:th_geomofS}, $\tilde{\bf a}_{(i)}^{\tsp} {\bf a}_{(j)} = \tilde{\bf a}_{(i)}^{\tsp}\frac{1}{\sqrt{2}}({\bf e}_{j} - {\bf e}_{j+1}) = \frac{1}{\sqrt{2}}(\tilde{\bf a}_{(i)j}^{\tsp}-\tilde{\bf a}_{(i)j+1}^{\tsp}) = 0$. Therefore $\tilde{\bf a}_{(i)j} = \tilde{\bf a}_{(i)j+1}$. Hence $\tilde{\bf a}_{(i)0} = \tilde{\bf a}_{(i)1}=\ldots=\tilde{\bf a}_{(i)i}$ and $\tilde{\bf a}_{(i)i+1} = \tilde{\bf a}_{(i)i+2}\ldots=\tilde{\bf a}_{(i)Q} = 0$. Which results in $\tilde{\bf a}_{(i)} = \frac{1}{\sqrt{i+1}} \sum_{l \in \infsetstyle{Z}_{i+1}} {\bf e}_{l}$. 
\item $i \in \{ Q+1, Q+2, \ldots, m-1 \}$ and $j \in \{ Q, Q+1, \ldots, m-1 \}\textbackslash \{i\}$: likewise, $\tilde{\bf a}_{(i)}^{\tsp} {\bf a}_{(j)} = \tilde{\bf a}_{(i)}^{\tsp}{\bf e}_{j} = \tilde{\bf a}_{(i)j}^{\tsp} = 0$. Hence $\tilde{\bf a}_{(i)Q} = \tilde{\bf a}_{(i)Q+1}=\ldots=\tilde{\bf a}_{(i)m-1} = 0$, excluding $\tilde{\bf a}_{(i)i}$, where $\tilde{\bf a}_{(i)i} \neq 0$ since $j\neq i$.
\item $i \in \{ Q+1, Q+2, \ldots, m-1 \}$ and $j \in \{ 0, 1, \ldots, Q-1 \}\textbackslash \{i\}$:  we know $\tilde{\bf a}_{(i)}^{\tsp} {\bf a}_{(j)} = \tilde{\bf a}_{(i)}^{\tsp} \frac{1}{\sqrt{2}}({\bf e}_{j} - {\bf e}_{j+1}) = 0$. Likewise $\tilde{\bf a}_{(i)0} = \tilde{\bf a}_{(i)1} = \ldots = \tilde{\bf a}_{(i)Q} = 0$.
\item $i=Q$ and $j \in \{Q+1, \ldots, m-1 \}$: Since $i\neq j$, then $j \in \{ Q+1, Q+2, \ldots, m-1 \}$. Hence $\tilde{\bf a}_{(i)}^{\tsp} {\bf a}_{(j)} = \tilde{\bf a}_{(i)}^{\tsp}{\bf e}_{j} = \tilde{\bf a}_{(i)j} = 0$. Therefore $\tilde{\bf a}_{(i)Q+1} = \ldots = \tilde{\bf a}_{(i)m-1} = 0$.
\item $i=Q$ and $j \in \{ 0, 1, \ldots, Q-1\}$: By defining $i=Q$, we know that $i\neq j$, hence $\tilde{\bf a}_{(i)}^{\tsp} {\bf a}_{(j)} = \tilde{\bf a}_{(i)}^{\tsp}\frac{1}{\sqrt{2}}({\bf e}_{j} - {\bf e}_{j+1}) = \frac{1}{\sqrt{2}}(\tilde{\bf a}_{(i)j}^{\tsp}-\tilde{\bf a}_{(i)j+1}^{\tsp}) = 0$, which implies $\tilde{\bf a}_{(i)0} = \tilde{\bf a}_{(i)1}=\ldots=\tilde{\bf a}_{(i)Q}$, or likewise $\tilde{\bf a}_{(i)} = \frac{1}{\sqrt{Q+1}} \sum_{l \in \infsetstyle{Z}_{Q+1}} {\bf e}_{l}$.
\end{enumerate}
\end{proof}\\

\begin{cor_geomofSII} 
 Let $\ordsetstyle{I} = (0, 1, \ldots,Q| Q \in \infsetstyle{Z}_m \textbackslash \{0\})$ be an (ordered) 
 preference tuple. Define $\infsetstyle{S}_{\ordsetstyle{I}}^\perp$ as per 
 definition $1$.  Using the notation of theorem 
 \ref{th:th_geomofS},  ${\bf v} \in \infsetstyle{S}_{\ordsetstyle{I}}^\perp$ if 
 and only if ${\bf v} = {\bf 0}$ or $\exists i \ne k \in \infsetstyle{Z}_m$ such 
 that $\sgn ( \tilde{\bf a}_{(i)}^{\tsp} {\bf v} ) \ne \sgn ( \tilde{\bf 
 a}_{(k)}^{\tsp} {\bf v} )$, where $\sgn (0) = 0$.
 \label{cor:cor_geomofSII}
\end{cor_geomofSII}
\begin{proof}
By definition of $\infsetstyle{S}_{\ordsetstyle{I}}^\perp$, ${\bf v} \in 
\infsetstyle{S}_{\ordsetstyle{I}}^\perp$ if $\exists {\bf s} \in 
\infsetstyle{S}_{\ordsetstyle{I}}$ such that ${\bf s}^{\tsp} {\bf v} = 0$.  
This is trivially true of ${\bf v} = {\bf 0}$.  Otherwise, using the generative 
representation of $\infsetstyle{S}_{\ordsetstyle{I}}$, there must exist ${\bf 
c} \in \posquad \backslash \{ {\bf 0} \}$ such that ${\bf s} = \sum_i c_i 
\tilde{\bf a}_{(i)}$.  Hence ${\bf v} \in \infsetstyle{S}_{\ordsetstyle{I}}^\perp 
\backslash \{ {\bf 0} \}$ only if there exists ${\bf c} \in \posquad \backslash 
\{ {\bf 0} \}$ such that ${\bf s}^{\tsp} {\bf v} = \sum_i c_i ({\bf 
a}_{(i)}^{\tsp} {\bf v}) = 0$ which, as ${\bf c} \ne {\bf 0}$, is only possible 
if $\exists i,k \in \infsetstyle{Z}_m$ such that $\sgn ( \tilde{\bf 
a}_{(i)}^{\tsp} {\bf v} ) \ne \sgn ( \tilde{\bf a}_{(k)}^{\tsp} {\bf v} ).$
\end{proof}

\section{Experiments}
\subsection{Poloni's two objective function}
\begin{figure*}[t]
\centering
\label{fig:Mould}
\begin{subfigure}[t]{0.31\linewidth}
    \centering
    \includegraphics[width=1\linewidth]{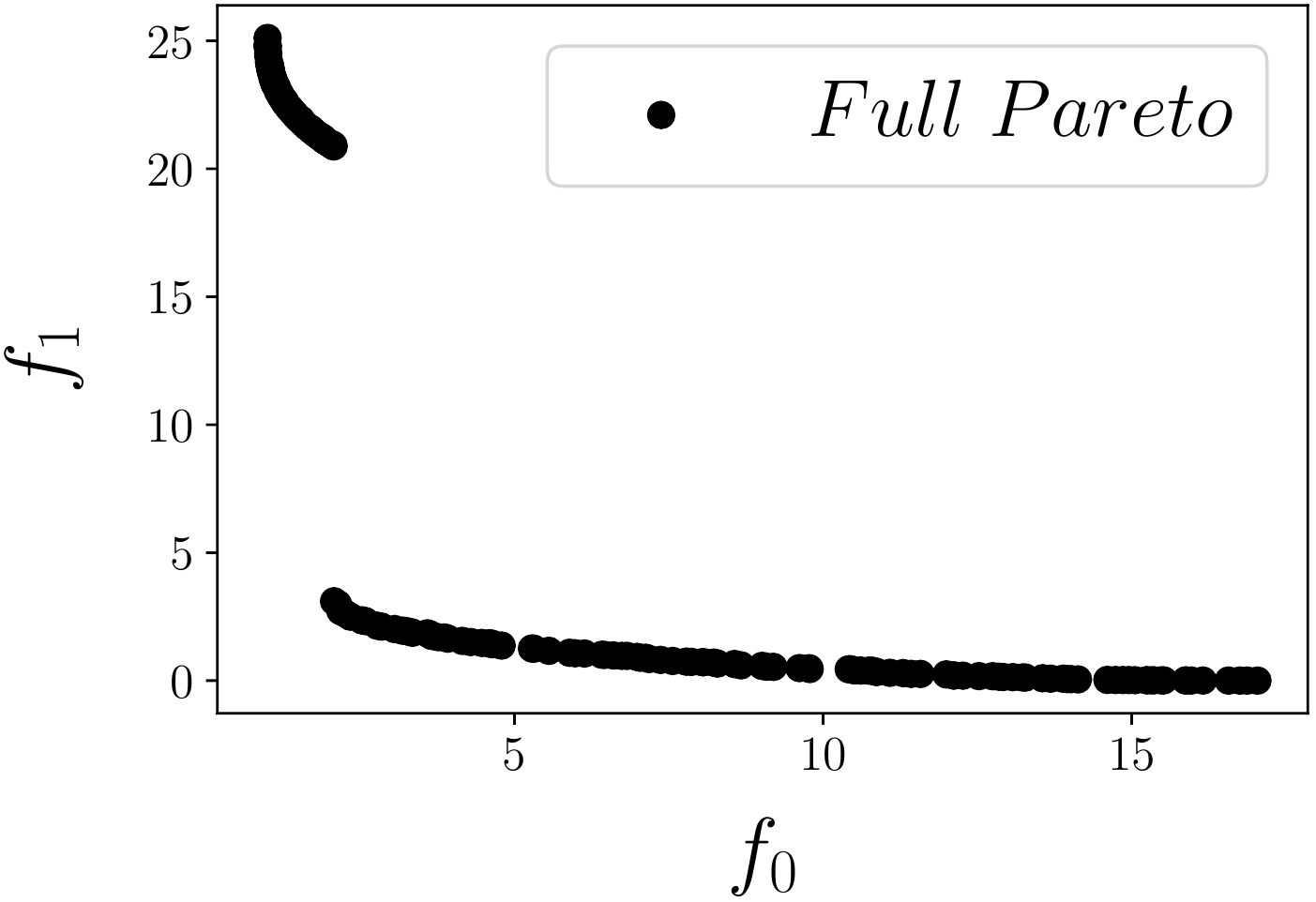}
    \caption{Full Pareto front}
    \label{fig:SP0}
\end{subfigure}
\begin{subfigure}[t]{0.31\linewidth}
    \centering
    \includegraphics[width=1\linewidth]{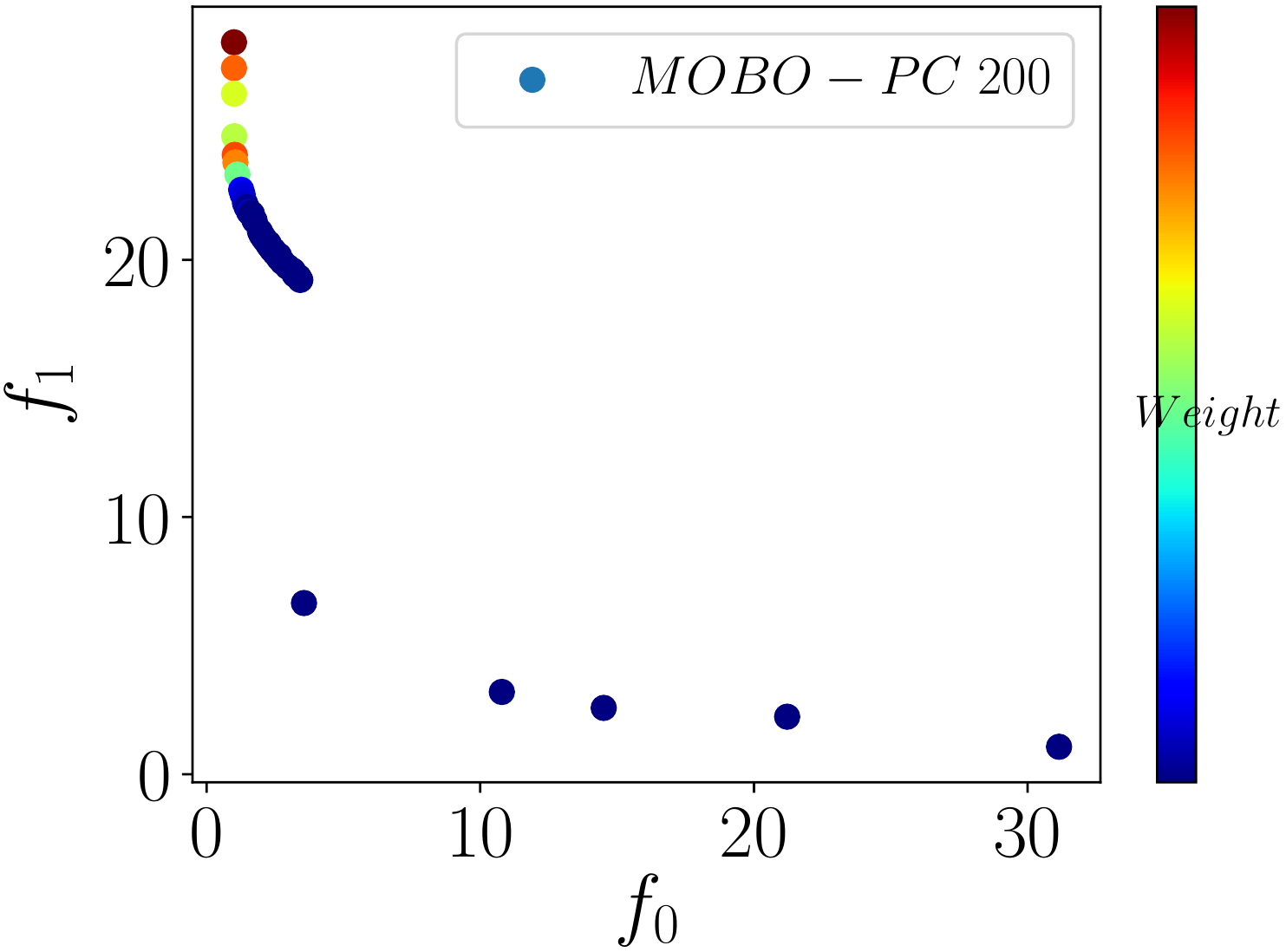}
    \caption{$s_0 > s_1$ preference-order constraint}
    \label{fig:SP1}     
\end{subfigure}
\begin{subfigure}[t]{0.31\linewidth}
    \centering
    \includegraphics[width=1\linewidth]{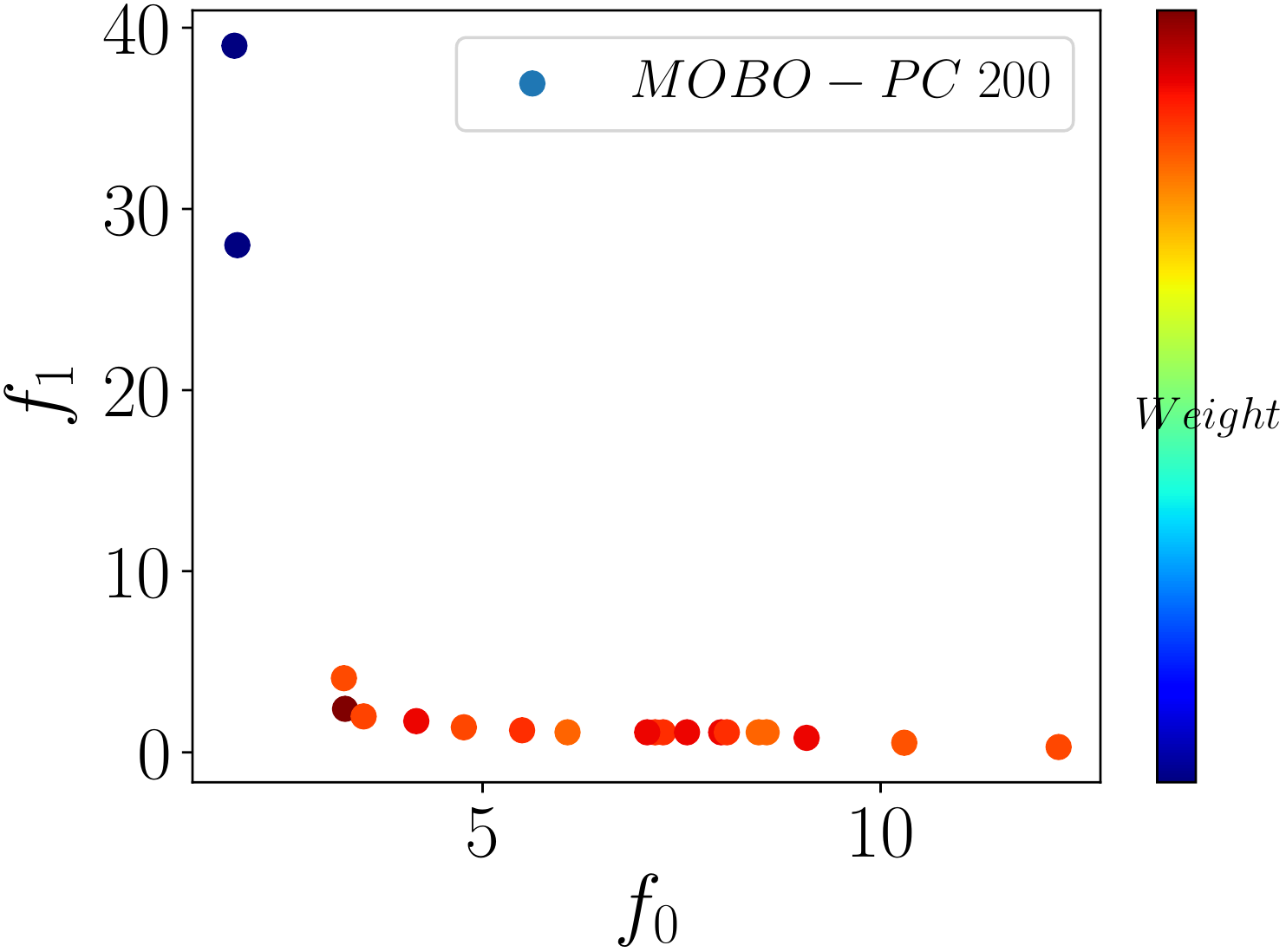}
    \caption{$s_1 > s_0$ preference-order constraint}
    \label{fig:SP2}
\end{subfigure}
\caption{Obtained solutions for Poloni's two objective function. \ref{fig:SP0} shows the full Pareto front. \ref{fig:SP1} illustrates the obtained solutions with $s_0 > s_1$ preference-order constraint on stability. And \ref{fig:SP2} shows the results of $s_1 > s_0$ or more stable solutions for $f_1$.}  
\label{fig:3d}
\end{figure*}

The results of our algorithm on Poloni's two objective function  \cite{deb2002scalable}. Figure \ref{fig:SP1} shows more stable results for $f_0$ than $f_1$. Likewise, in figure \ref{fig:SP2}   stability of solutions in $f_1$ is favored over $f_0$.
\begin{figure*}[t]
\centering
\label{fig:Mould}
\begin{subfigure}[t]{0.31\linewidth}
    \centering
    \includegraphics[width=1\linewidth]{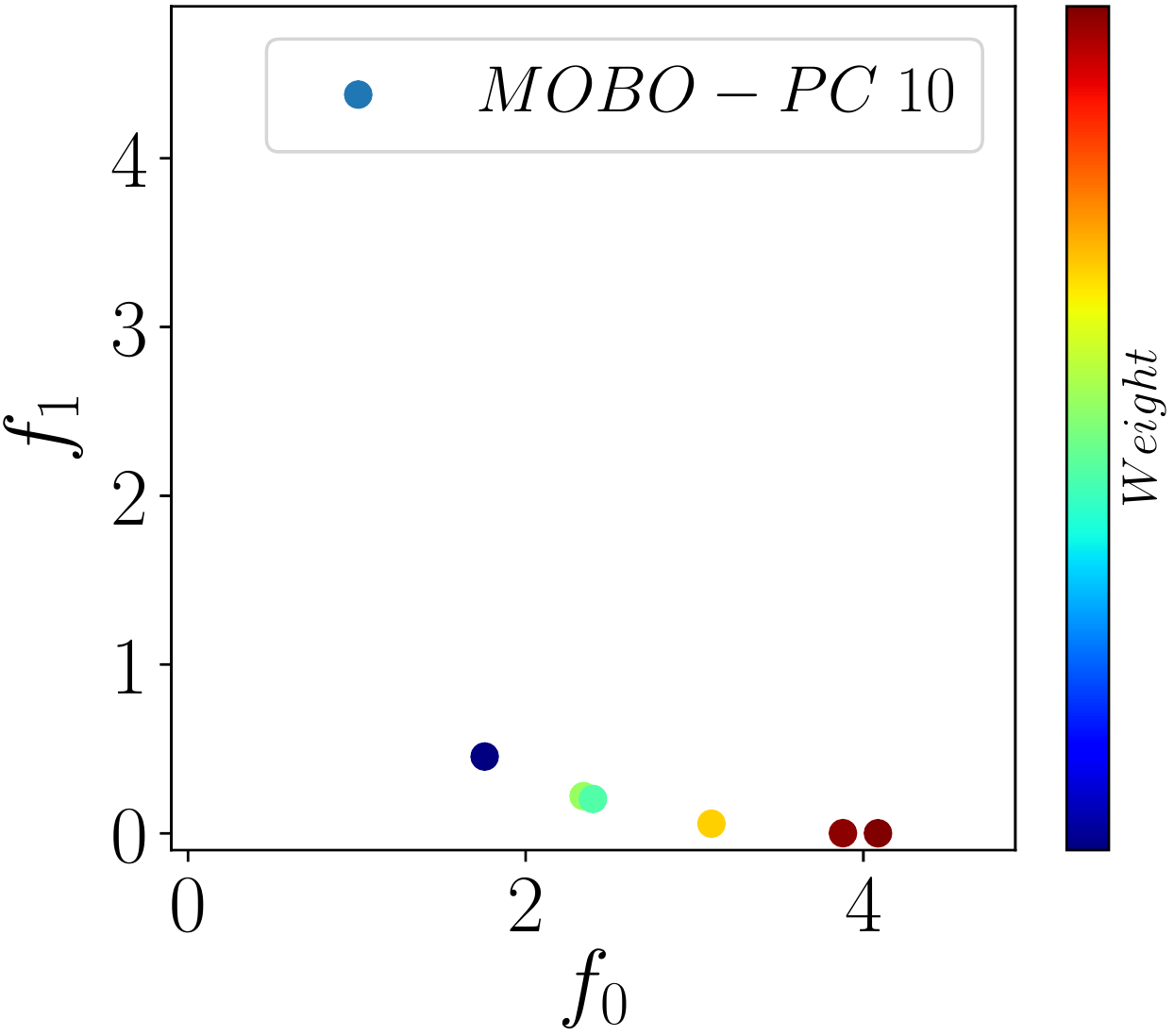}
    \caption{Iteration number $10$}
    \label{fig:SP_d2}
\end{subfigure}
\begin{subfigure}[t]{0.31\linewidth}
    \centering
    \includegraphics[width=1\linewidth]{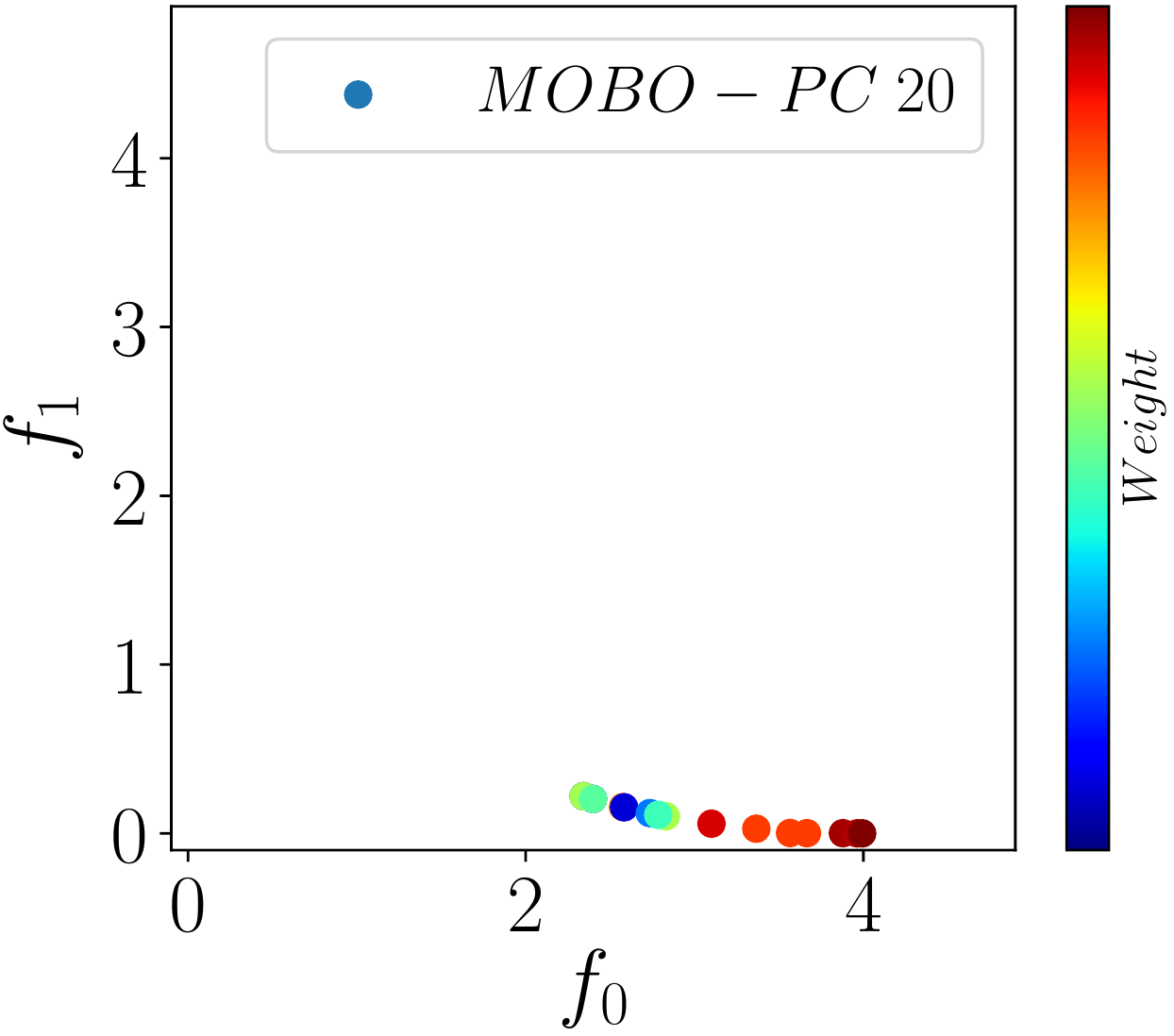}
    \caption{Iteration number $20$}
    \label{fig:SP_d3}
\end{subfigure}
\begin{subfigure}[t]{0.31\linewidth}
    \centering
    \includegraphics[width=1\linewidth]{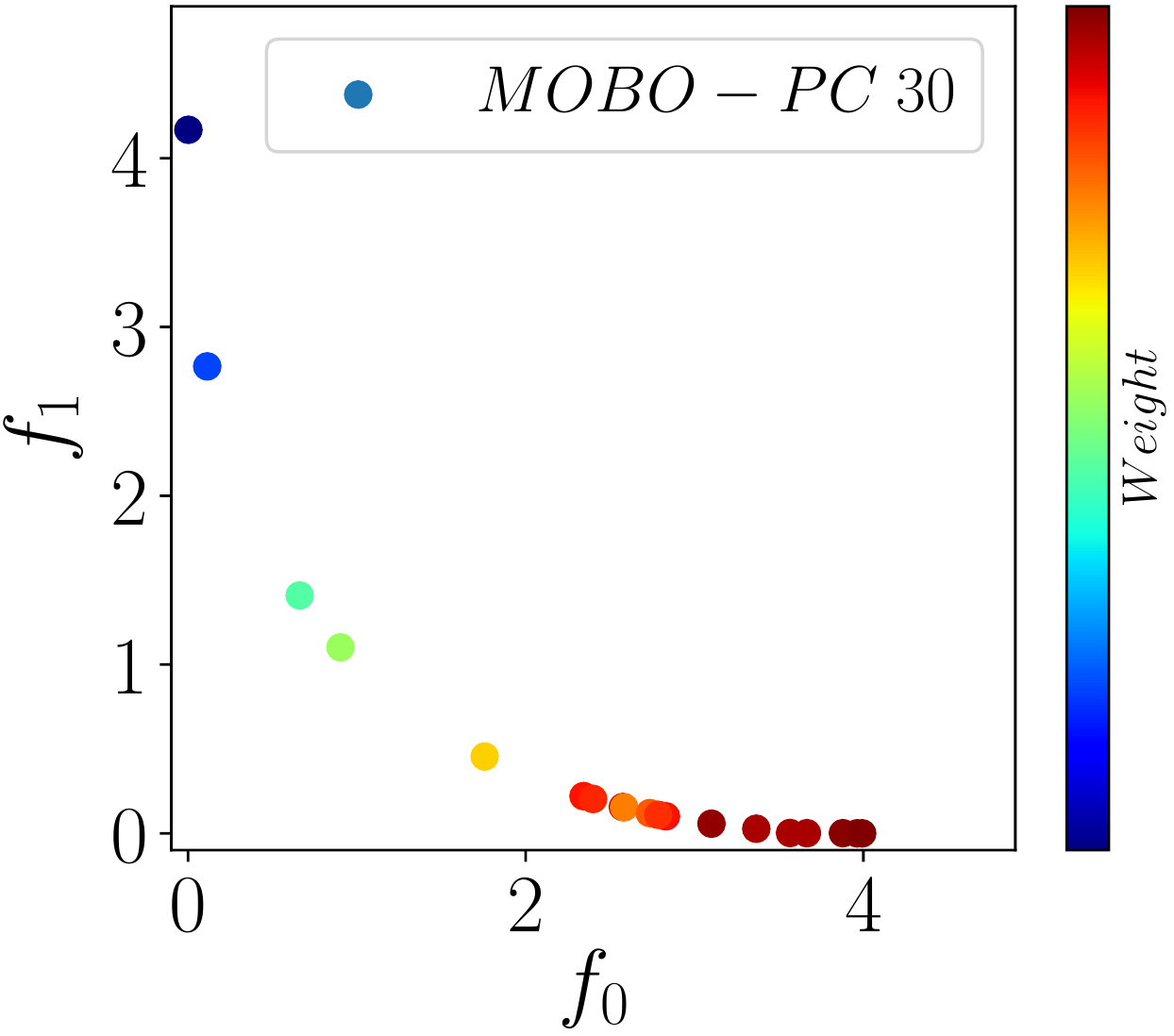}
    \caption{Iteration number $30$}
    \label{fig:SP_d4}     
\end{subfigure}\\
\begin{subfigure}[t]{0.31\linewidth}
    \centering
    \includegraphics[width=1\linewidth]{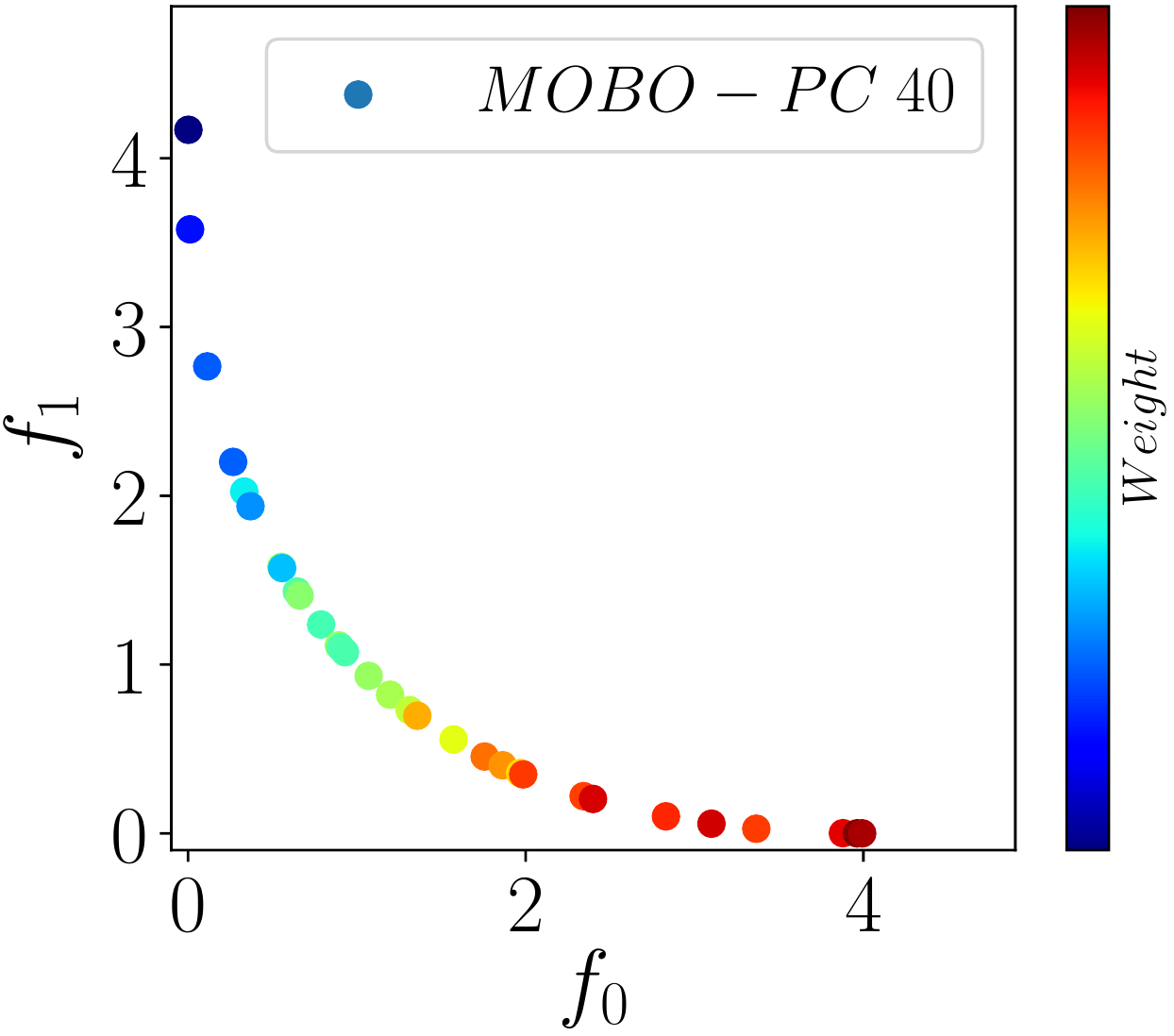}
    \caption{Iteration number $40$}
    \label{fig:SP_d5}
\end{subfigure}
\begin{subfigure}[t]{0.31\linewidth}
    \centering
    \includegraphics[width=1\linewidth]{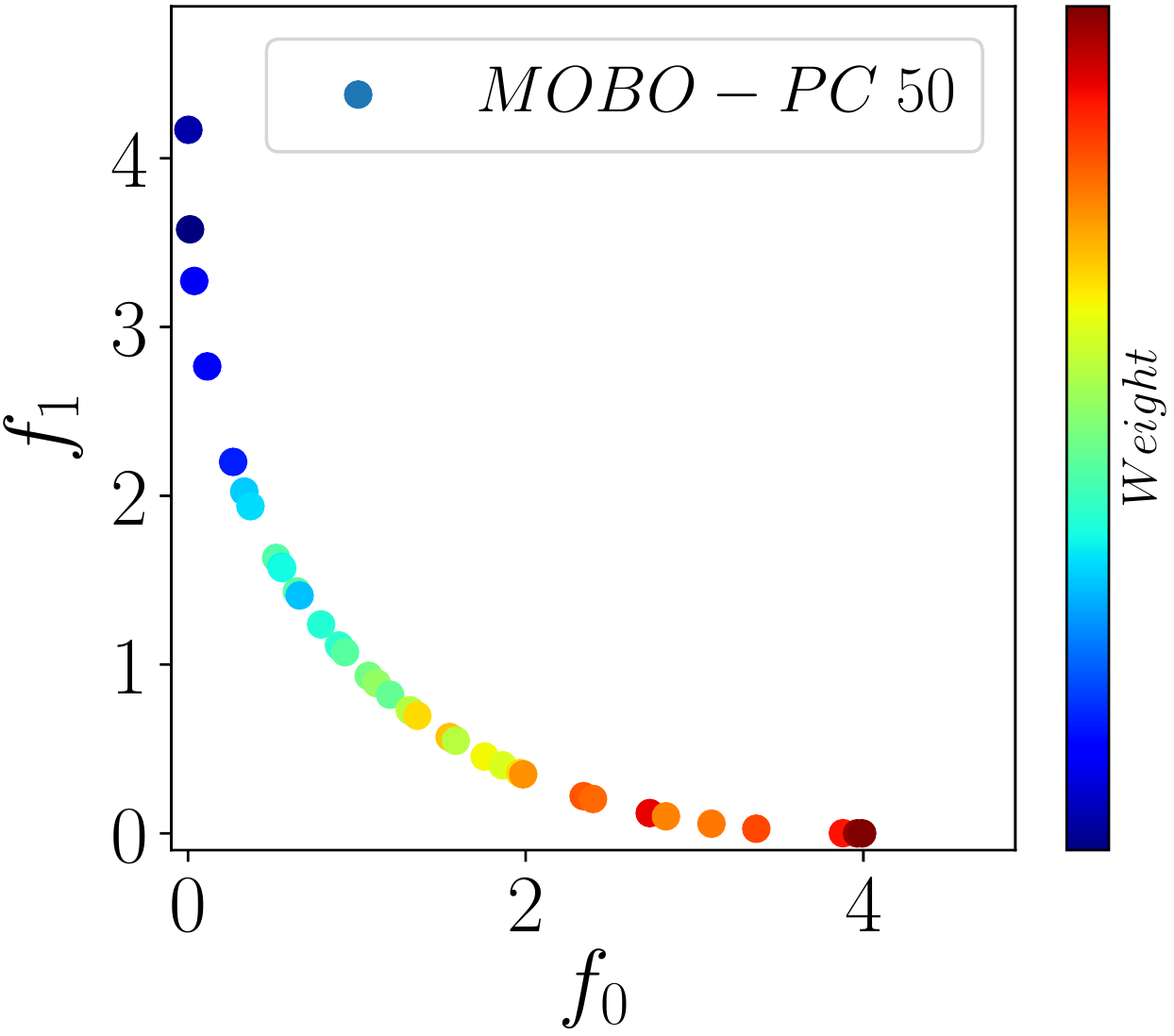}
    \caption{Iteration number $50$}
    \label{fig:SP_d6}     
\end{subfigure}
\begin{subfigure}[t]{0.31\linewidth}
    \centering
    \includegraphics[width=1\linewidth]{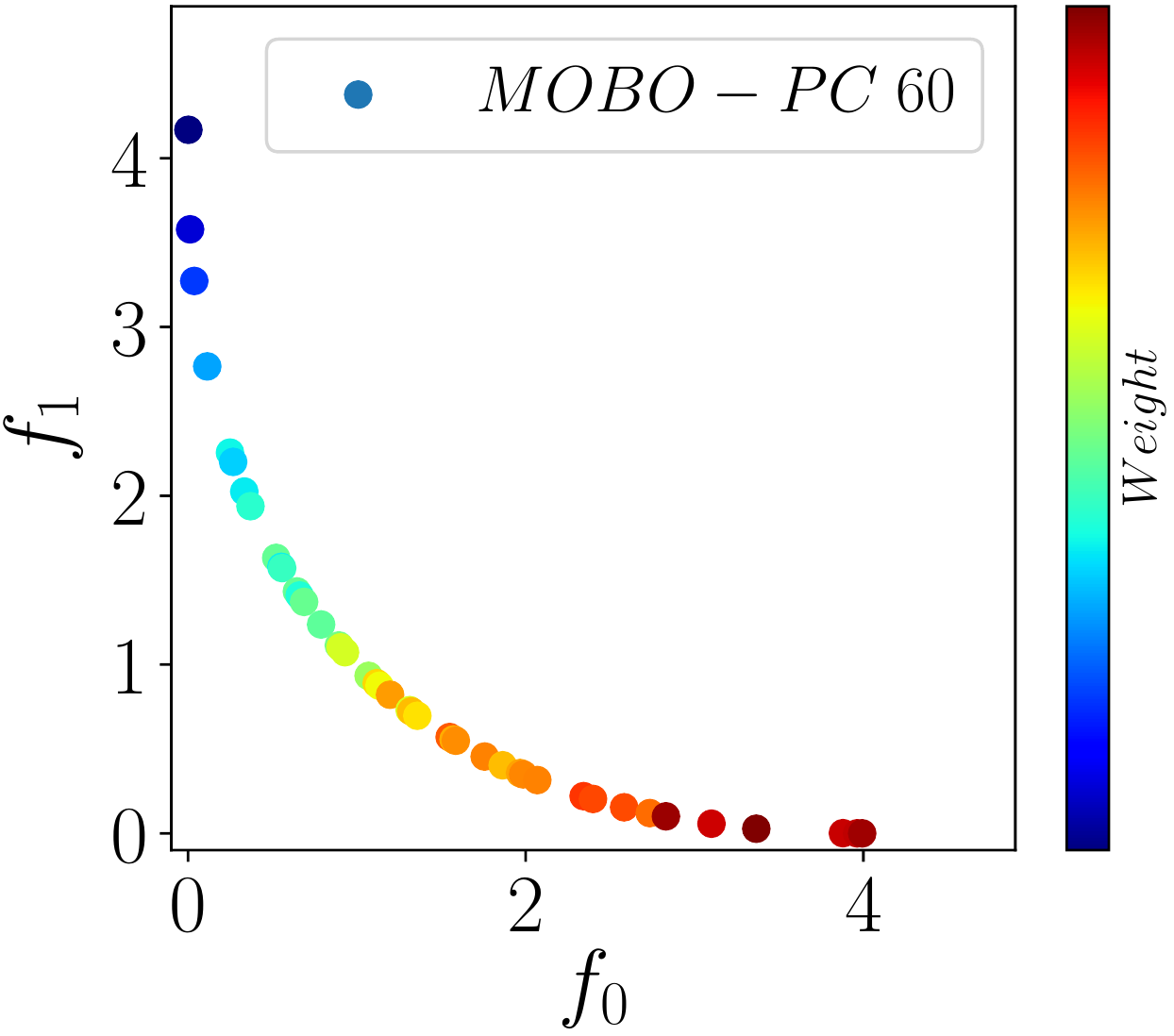}
    \caption{Iteration number $60$}
    \label{fig:SP_d7}
\end{subfigure}
\caption{Illustration of the progress of MOBO-PC on Schaffer function N. $1$.}  
\label{fig:SP_d1}
\end{figure*}

\subsection{Progress of MOBO-PC}
The calculation of hypervolume in MOBO-PC relies on both values of the weights for the Pareto front points and also the improved volume. That will result in favoring solutions complying with the constraints and assign them higher weights comparing to other Pareto front solutions. But if the amount of volume to be improved is insignificant, the acquisition function favors solutions with lesser compliance with constraints but more possibility to increase the amount of volume -i.e.  the region with most compliance with constraints is well explored and occupied with many Pareto front solutions, so the amount of hypervolume improvement drops due to the small amount of improvement in the volume despite of higher weights for the solutions in that region. Hence, the algorithm will then look for more diverse solutions that can increase the amount of hypervolume, that is the solutions which are more diverse and less stable. Figure \ref{fig:SP_d1} illustrates how acquisition function favors the points with higher weights at the first, and then lean towards the more unexplored regions with higher amount of hypervolume improvement.

\begin{figure*}[t]
\centering
\label{fig:Mould}
\begin{subfigure}[t]{0.35\linewidth}
    \centering
    \includegraphics[width=1\linewidth]{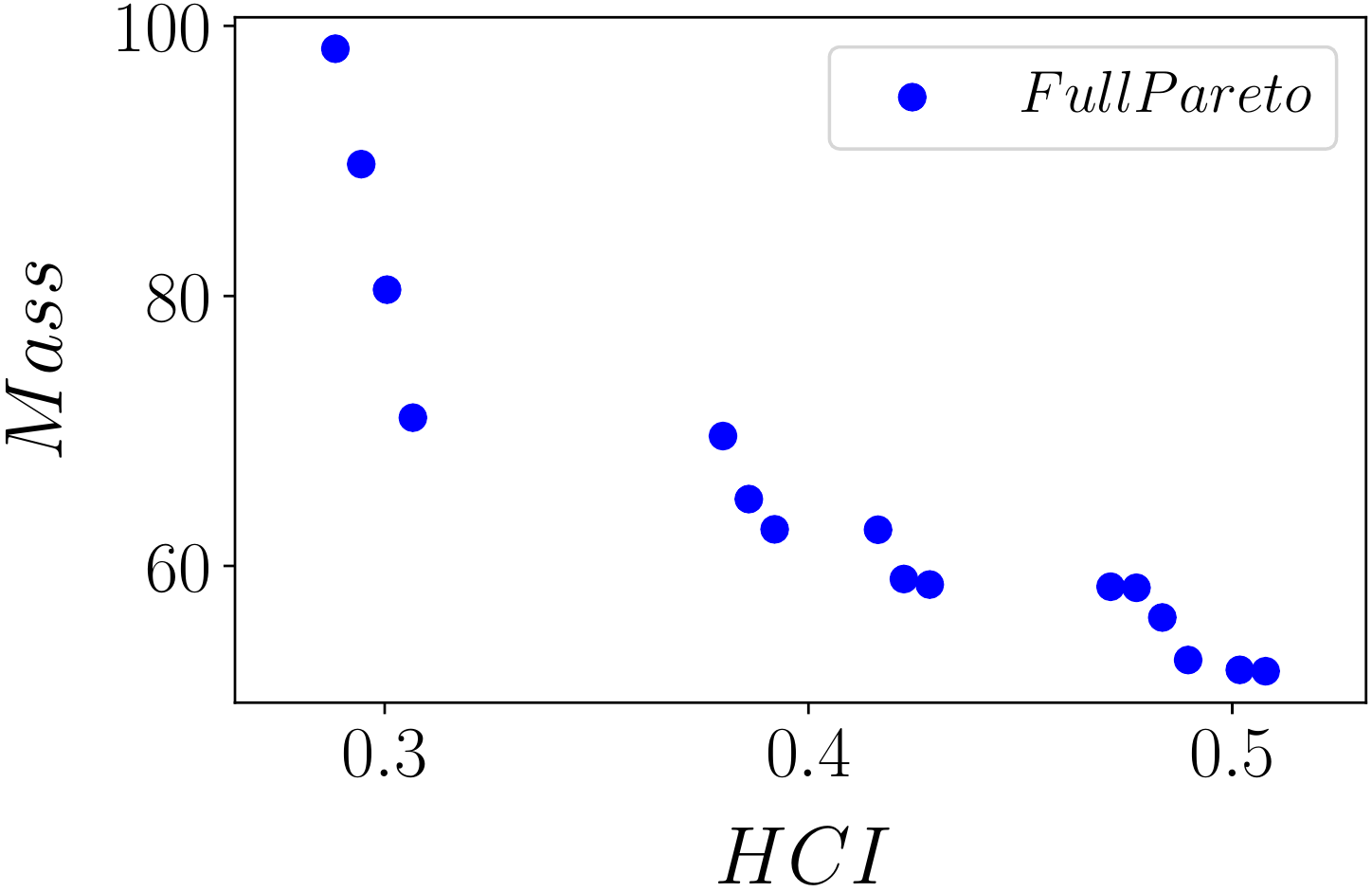}
    \caption{Full Pareto front}
    \label{fig:sp4}
\end{subfigure}
\begin{subfigure}[t]{0.31\linewidth}
    \centering
    \includegraphics[width=1\linewidth]{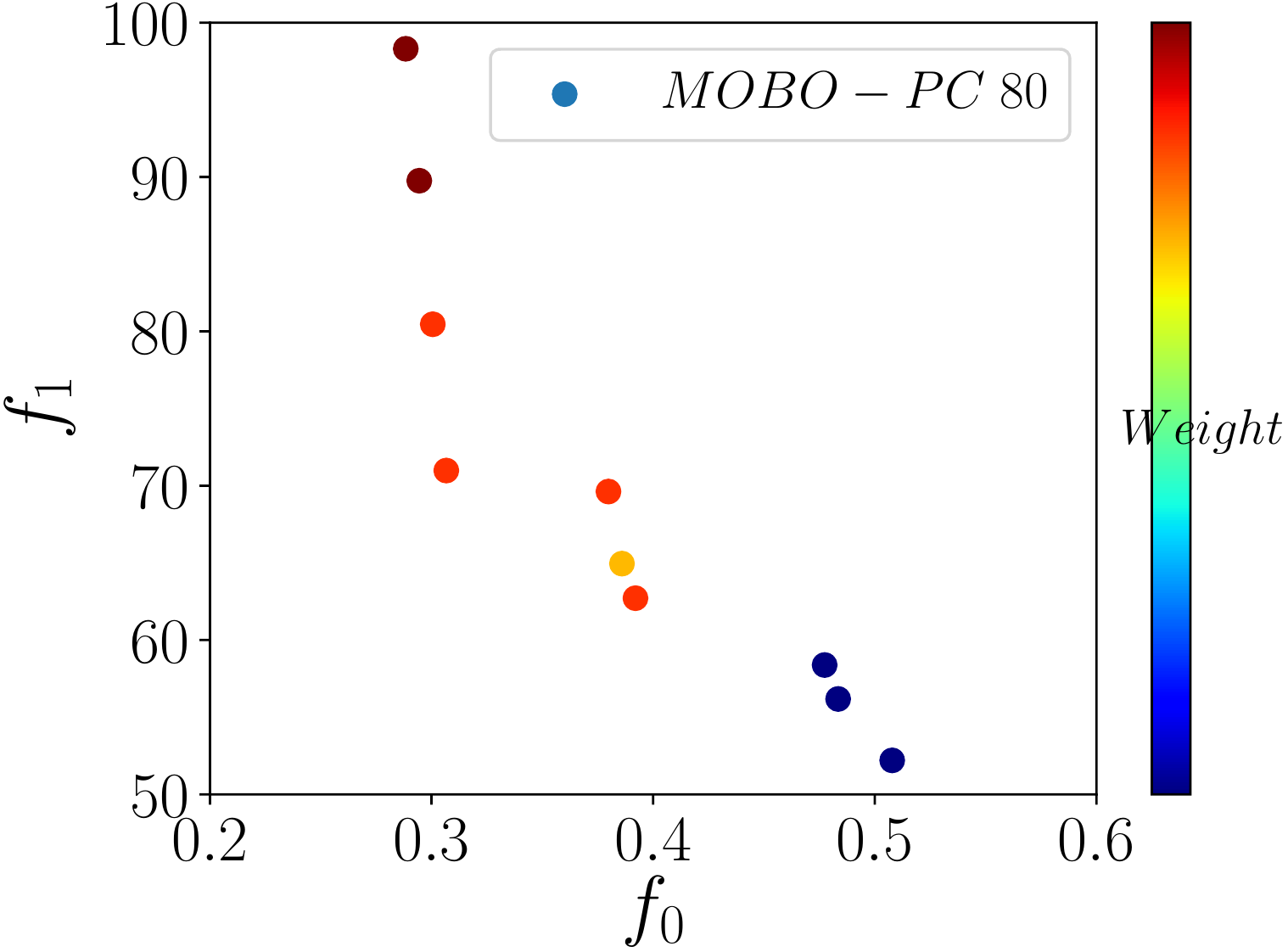}
    \caption{$s_0 > s_1$ preference-order constraint}
    \label{fig:sp5}     
\end{subfigure}
\begin{subfigure}[t]{0.31\linewidth}
    \centering
    \includegraphics[width=1\linewidth]{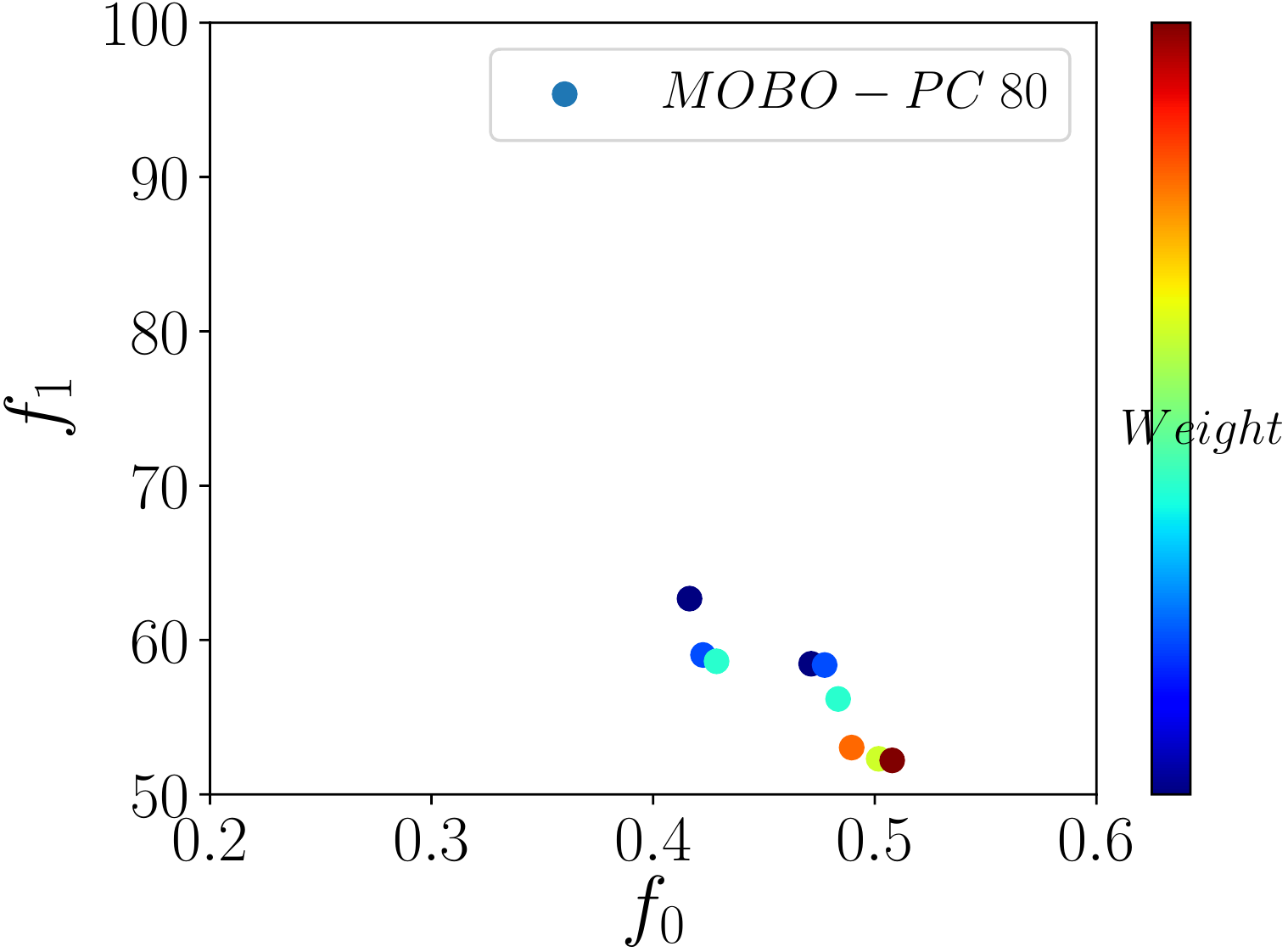}
    \caption{$s_1 > s_0$ preference-order constraint}
    \label{fig:sp6}
\end{subfigure}
\caption{Obtained solutions for Simple Crash Study.}  
\label{fig:4sp}
\end{figure*}

\section{Simple Crash Study}
\subsubsection{Summary}
This problem concerns a collision in which a simplified vehicle moving at constant velocity crashes into a pole. The input parameters vary the strength of the bumper and hood of the car. During the crash, the front portion of the car deforms. The design goal is to maximise the \textit{crashworthiness} of the vehicle. If the car is too rigid, the passenger experiences injury due to excessive forces during the impact. If the car is not rigid enough, the passenger may be crushed as the front of the car intrudes into the passenger space. This dataset is available in https://bit.ly/2FWNXQS.

\subsubsection{Input Variables}
\begin{itemize}
\item tbumper is the mass of the front bumper bar. Range of tbumper is between $1$ and $5$.
\item thood is the mass of the front, hood and underside of the bonnet. Range of thood is between $1$ and $5$.
\end{itemize}

\subsubsection{Output Responses}
\begin{itemize}
\item Intrusion is the intrusion of the car frame into the passenger space. This is computed from the change in the separation of two points, one on the front of the car (Node $\#167$) and one of the roof (Node $\#432$). Lower intrusion is better. Increasing the mass of the hood and bumper will reduce the intrusion.
\item HIC is the head injury coefficient. This is computed from the maximum deceleration during the collision. Lower HIC is better. Increasing the mass of the hood and bumper will increase the HIC.
\item Mass is the combined mass of the front structural components. Lower mass is better.
\end{itemize}
In our experiment, we are using HIC as $f_0$ and Mass as $f_1$ to be minimised simultaneously.
\par 
Figure \ref{fig:4sp} shows the obtained results for this problem. Figure \ref{fig:sp4} illustrates full Pareto front, as \ref{fig:sp5} and \ref{fig:sp6} demonstrates our obtained results based on the defined preference-order constraints. Figure \ref{fig:sp5} shows that the Pareto front points are more stable in $f_0$ (HIC) than $f_1$ (Mass). As for figure \ref{fig:sp6} Pareto front solutions are in favor of stability for $f_1$ and more diversity for $f_0$.
\end{document}